\documentclass[sigconf]{acmart} 
\usepackage{listings}
\usepackage{times}  
\usepackage{helvet}  
\usepackage{courier}  
\usepackage{xr}

\usepackage[utf8]{inputenc}  
\usepackage{CJKutf8}  

\usepackage{bm}
\usepackage{caption} 
\usepackage{algorithm}
\usepackage{algorithmic}

\usepackage{newfloat}
\DeclareCaptionStyle{ruled}{labelfont=normalfont,labelsep=colon,strut=off} 
\floatstyle{ruled}
\newfloat{listing}{tb}{lst}{}
\floatname{listing}{Listing}

\usepackage{color}
\usepackage{pgfplots}
\usepackage{multirow}
\usepackage{tabularx}
\usepackage{calligra}
\usepackage{amsmath}
\usepackage{listings}
\usepackage{adjustbox}
\usepackage{pgf-pie}
\usetikzlibrary{pgfplots.polar}

\usepackage{booktabs}

\usepackage{subcaption}
\usepackage{graphicx}

\definecolor{color1}{RGB}{159,168,218}
\definecolor{color2}{RGB}{173,201,198}
\definecolor{color3}{RGB}{207,216,220}
\definecolor{color4}{RGB}{247,194,183}
\definecolor{color5}{RGB}{187,230,190}
\definecolor{color6}{RGB}{187,230,222}
\definecolor{darkgreen}{rgb}{0,0.5,0}
\definecolor{fancy_color1}{RGB}{51,75,127}
\definecolor{fancy_color2}{RGB}{180,75,144}
\definecolor{pie_pink}{HTML}{feaba7}
\definecolor{pie_rose}{HTML}{fbe1ed}
\definecolor{pie_purple}{HTML}{dbd1ed}
\definecolor{pie_green}{HTML}{adeaad}
\definecolor{pie_orange}{HTML}{ffcd88}
\definecolor{pie_gray}{HTML}{dde7eb}
\definecolor{orchid}{RGB}{218, 112, 214}

\definecolor{radar_color1}{HTML}{0077ad}
\definecolor{radar_color2}{HTML}{aa71c7}
\definecolor{radar_color3}{HTML}{ff7f4f}
\definecolor{radar_color4}{HTML}{e765af}
\definecolor{radar_color5}{HTML}{6178c5}
\definecolor{radar_color6}{HTML}{ffa600}
\definecolor{radar_color7}{HTML}{ff6684}

\definecolor{bar_color1}{HTML}{0099de}
\definecolor{bar_color2}{HTML}{7a92ee}
\definecolor{bar_color3}{HTML}{c684e5}
\definecolor{bar_color4}{HTML}{ff72c2}
\definecolor{bar_color5}{HTML}{ff6e8d}
\definecolor{bar_color6}{HTML}{ff8453}
\definecolor{bar_color7}{HTML}{ffa600}

\setcopyright{acmlicensed}
\copyrightyear{2025}
\acmYear{2025}
\acmDOI{TBD}

\acmConference[WWW '25, NLP4KGC]{}{April 28-May 2, 2025}{Sydney, NSW, Australia}
\acmISBN{TBD}

\begin{document}

\title[Graphusion]{Graphusion: A RAG Framework for Scientific Knowledge Graph Construction with a Global Perspective}




\author{Rui Yang}
\affiliation{%
  \institution{Duke-NUS Medical School}
  \country{Singapore}
}
\email{yang_rui@u.nus.edu}

\author{Boming Yang}
\affiliation{%
  \institution{The University of Tokyo}
  \country{Tokyo, Japan}}
\email{boming@g.ecc.u-tokyo.ac.jp}

\author{Xinjie Zhao}
\affiliation{%
    \institution{The University of Tokyo}
    \country{Tokyo, Japan}}
\email{xinjie-zhao@g.ecc.u-tokyo.ac.jp}

\author{Fan Gao}
\affiliation{%
    \institution{The University of Tokyo}
    \country{Tokyo, Japan}}
\email{fangao0802@gmail.com}

\author{Aosong Feng}
\affiliation{%
 \institution{Yale University}
 \country{New Haven, CT, USA}}
\email{aosong.feng@yale.edu}
\author{Sixun Ouyang}
\affiliation{%
  \institution{Smartor Inc.}
  \country{Shanghai, China}}
\email{troy.oysx@gmail.com}

\author{Moritz Blum}
\affiliation{%
  \institution{Bielefeld University}
  \country{Bielefeld, Germany}}
\email{mblum@techfak.uni-bielefeld.de}

\author{Tianwei She}
\affiliation{%
  \institution{Smartor Inc.}
  \country{Shanghai, China}}
\email{tianwei.v.she@gmail.com}

\author{Yuang Jiang}
\affiliation{%
  \institution{Smartor Inc.}
  \country{Shanghai, China}}
\email{jiangyuang1995@gmail.com}

\author{Freddy Lecue}
\affiliation{%
  \institution{INRIA}
  \country{Paris, France}}
\email{freddy.lecue@inria.fr}
  
\author{Jinghui Lu}
\affiliation{%
  \institution{Smartor Inc.}
  \country{Shanghai, China}}
\email{liuxiangtian213@gmail.com}

\author{Irene Li}
\affiliation{%
  \institution{University of Tokyo}
    \country{Tokyo, Japan}
}
\email{ireneli@ds.itc.u-tokyo.ac.jp}

\renewcommand{\shortauthors}{Rui Yang, Boming Yang, Xinjie Zhao, Fan Gao and et al.}

\begin{abstract}
Knowledge Graphs~(KGs) are crucial in the field of artificial intelligence and are widely used in downstream tasks, such as question-answering~(QA). The construction of KGs typically requires significant effort from domain experts. Large Language Models~(LLMs) have recently been used for Knowledge Graph Construction~(KGC). However, most existing approaches focus on a local perspective, extracting knowledge triplets from individual sentences or documents, missing a fusion process to combine the knowledge in a global KG. This work introduces Graphusion, a zero-shot KGC framework from free text. It contains three steps: in Step 1, we extract a list of seed entities using topic modeling to guide the final KG includes the most relevant entities; in Step 2, we conduct candidate triplet extraction using LLMs; in Step 3, we design the novel fusion module that provides a global view of the extracted knowledge, incorporating entity merging, conflict resolution, and novel triplet discovery. Results show that Graphusion achieves scores of 2.92 and 2.37 out of 3 for entity extraction and relation recognition, respectively. Moreover, we showcase how Graphusion could be applied to the Natural Language Processing~(NLP) domain and validate it in an educational scenario. Specifically, we introduce TutorQA, a new expert-verified benchmark for QA, comprising six tasks and a total of 1,200 QA pairs. Using the Graphusion-constructed KG, we achieve a significant improvement on the benchmark, for example, a 9.2\% accuracy improvement on sub-graph completion.
\end{abstract}




\maketitle



\section{Introduction}
Retrieval-Augmented Generation (RAG) \cite{Lewis2020RetrievalAugmentedGF} combines the advantages of retrieval methods and generative models, which improves the accuracy and relevance of generated content \cite{yang2024retrieval}. For instance, given a free-text corpus and a query involving two related entities, RAG can retrieve relevant information and infer their relation. Therefore, RAG can be used to enhance the performance of various knowledge-intensive tasks \cite{gao2024evaluating,yang-etal-2024-kg,yang2024ascle}. However, the need for more structured and comprehensive knowledge integration has highlighted the importance of adopting Knowledge Graphs (KGs). KGs provide structured and interconnected representations of information, offering richer context and reasoning capabilities that further enhance retrieval-augmented methods in complex applications. \cite{DBLP:conf/dl4kg/LiY23,bosselut-etal-2019-comet,DBLP:journals/tkde/PanLWCWW24} Especially in the scientific domain, KGs play a crucial role when precise extraction and modeling of entities and their relations are required. The accurate selection of entities with appropriate granularity, along with precise modeling of their relations, is key to capturing the complex structure and semantics inherent in scientific knowledge \cite{jiang2023graphcare,Li2021UnsupervisedCP}.


\begin{figure}[t]
    \centering
    \includegraphics[width=1.00\linewidth]{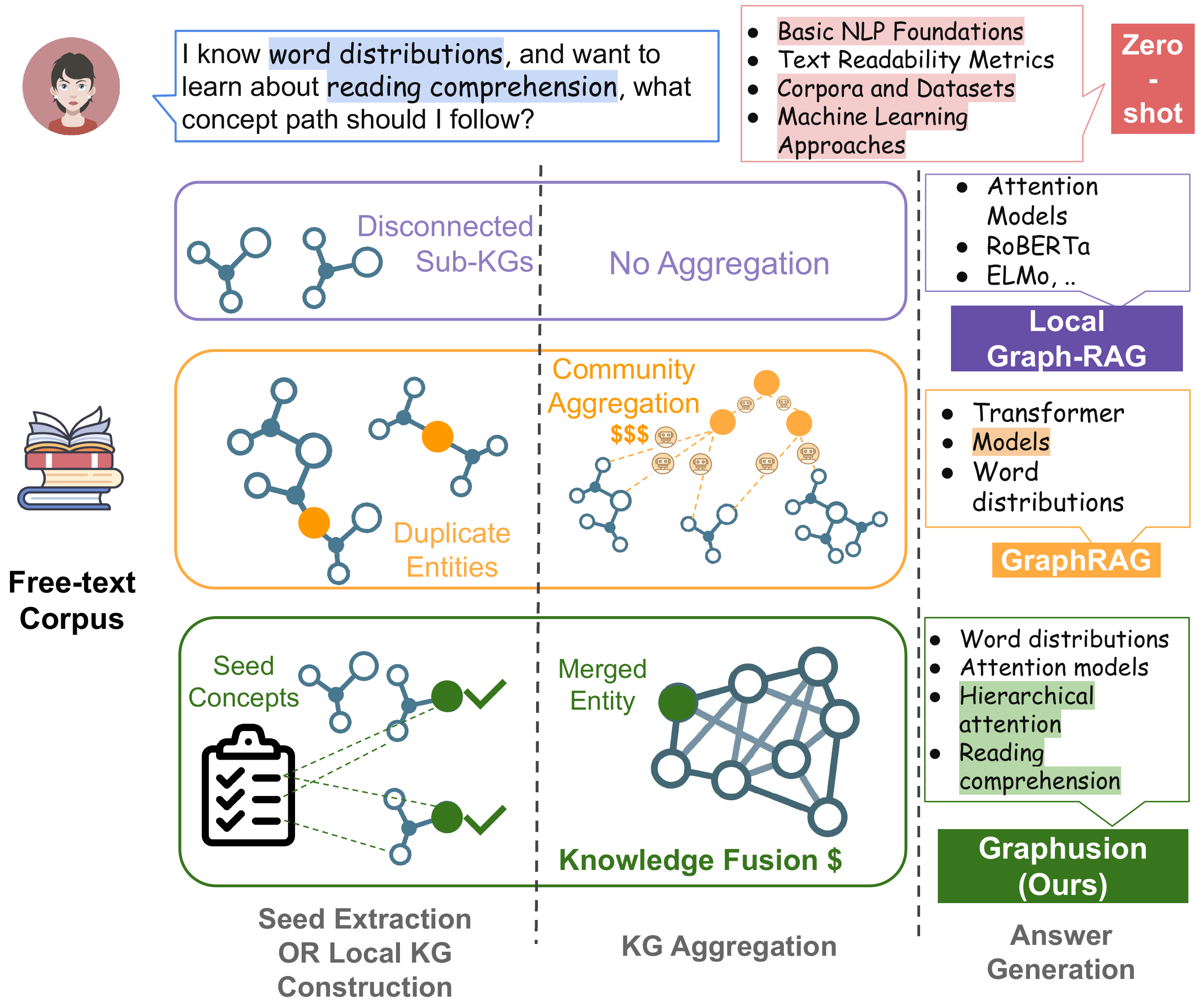}
    \caption{Comparison of Zero-shot LLM, RAG framework, and our Graphusion framework on applying LLMs for KGC.}
    \Description{Person interacting with the different agents, each implementing a framework.}
    \label{fig:intro}
    \vspace{-6mm}
\end{figure}

Existing knowledge graph construction (KGC) methods predominantly adopt a localized perspective ~\cite{Sheng2022ChallengingTA, Baek2023KnowledgeAugmentedLM, Carta2023IterativeZL}, focusing on extracting triplets at the sentence or paragraph level. While this approach works well for shallow knowledge—such as (\texttt{people, belong\_to, organization})—it falls short in scientific domains, where a global view is essential for identifying complex, multi-layered relations between entities. Localized methods often fail to capture the comprehensive and interconnected nature of knowledge, leading to limited accuracy and completeness when triplets are derived from isolated text segments, which is particularly crucial for scientific KGs. The recent success of GraphRAG \cite{Edge2024FromLT} highlights the value of leveraging a global KG for query-based summarization with the help of large language models (LLMs) \cite{achiam2023gpt}. However, despite offering enhanced contextual understanding, building and maintaining large-scale graph structures with hierarchical clustering significantly increases computational cost and complexity compared to simpler retrieval methods, limiting its application in resource-constrained environments or real-time systems requiring low-latency responses. Moreover, its effectiveness in scientific KGC, particularly when high entity granularity is required, remains unclear.

\begin{figure*}[t!]
    \centering
    \makebox[\textwidth][c]{
        \includegraphics[width=0.99\textwidth]{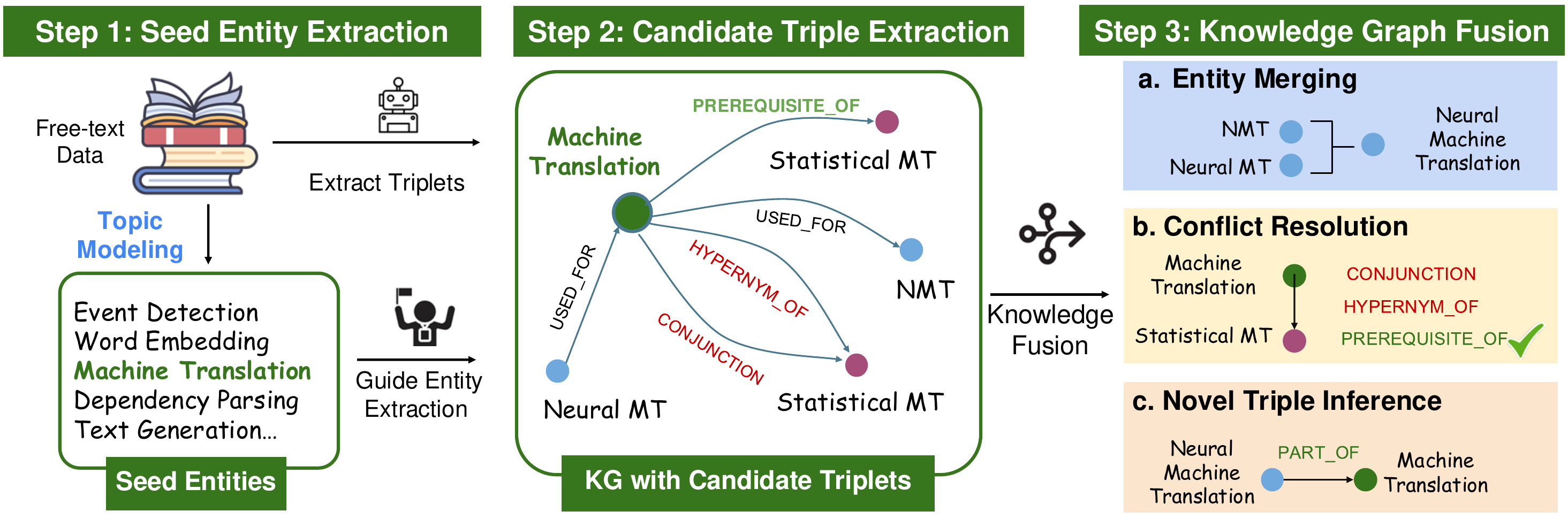}
    }
    \caption{Graphusion framework illustration. Graphusion consists of 3 steps: S1 Seed Entity Generation, S2 Candidate Triplet Extraction and S3 Knowledge Graph Fusion.}
    \Description{Conceptual illustration of the Graphuson framework.}
    \label{fig:model}
    \vspace{-3mm}
\end{figure*}

We illustrate the necessity of balancing the performance-efficiency trade-off in building the large-scale KG tailored to a user prompt in Fig.~\ref{fig:intro}. A user poses a specific question from the NLP domain, requiring a learning path from the entity \texttt{word distributions} to the entity \texttt{reading comprehension}. Ideally, the model should understand the relations between these entities and effectively map out the learning path. Zero-shot LLMs might provide somewhat relevant answers but tend to be too general, as shown in the figure, offering broad entities like \texttt{Basic NLP Foundations} or introducing confusing, inaccurately specific entities (e.g., \texttt{Corpora and Datasets}). While some RAG-based methods for building KGs focus primarily on relation extraction from limited sources via retrieval, this often results in numerous disconnected KGs or sub-graphs, we marked this method as Local Graph-RAG. We argue that a global understanding of domain knowledge is crucial and may be challenging for typical existing RAG frameworks. For instance, determining the relation between \texttt{hierarchical attention network} and \texttt{reading comprehension} may be difficult, as these entities might not appear within the same document. The model needs to extract and synthesize information from two (or more) documents to recognize that the relation is \texttt{Used\_for}. GraphRAG addresses this challenge hierarchically by linking entities to gather information on a global scale. However, despite the method's high cost, it remains unclear how relation conflicts are managed. Moreover, we observe that some entities, such as \texttt{models}, may be overly broad and therefore not useful for users.

Recognizing these limitations, we propose a cost-efficient KGC approach to incorporate the global perspective and improve RAG performance on downstream question-answering (QA) tasks.
Fig.~\ref{fig:intro} illustrates the differences between zero-shot LLMs and Local Graph-RAG, GraphRAG and our model. While Local Graph-RAG primarily focuses on knowledge extraction from a disconnected point of view, GraphRAG is able to summarize the information from a global view. However, our approach, Graphusion, incorporates a knowledge fusion step to integrate local knowledge into a global context directly. The core fusion step performs global merging and resolution across multiple local sub-graphs to form one large connected KG. Specifically, we leverage LLMs not only for extraction but also for critical knowledge integration, marking the first initiative to utilize LLMs for such a comprehensive merging process. Moreover, we generate a seed entity list to guide the entity extraction with better granularity. We demonstrate Graphusion's capability in knowledge graph construction and link prediction. Our results show that Graphusion achieves scores of 2.92 and 2.37 out of 3 for entity extraction and relation recognition, respectively. Furthermore, we show that a simplified link prediction prompt outperforms supervised learning baselines, achieving a 3\% higher F1 score. Then, to further show the power of the constructed KG, we showcase how it could be applied to the NLP domain, and most importantly, we validate it in an educational scenario with complex QA tasks. \footnote{\url{https://github.com/IreneZihuiLi/Graphusion}}

\section{Related Work}


\textbf{Knowledge Graph Construction} KGC aims to create a structured representation of knowledge in the form of a KG. A KG can be created from various sources, such as databases or texts. In our work, we focus on KGC from natural language resources. Research on KGs spans various domains, including medical, legal, and more~\cite{wikidata, Li2020RVGAERG, LeTuan2022TowardsBL, Kalla2023ScientificKG, ahrabian2023pubgraph}. Typically, KGC from text involves several methods such as entity extraction and link prediction~\cite{Luan2018MultiTaskIO,Reese2020KGCOVID19AF}, with a significant focus on supervised learning. Recently, LLMs have been utilized in KGC relying on their powerful zero-shot capabilities~\cite{Zhu2023LLMsFK, Carta2023IterativeZL,Zhang2023MakingLL,Li2024LLMbasedMK}. Although relevant works propose pipelines for extracting knowledge, they often remain limited to localized views, such as extracting triplets from the sentence or paragraph level. In contrast, our work focuses on shifting from a local perspective to a global one, aiming to generate a more comprehensive KG. Approaches such as GraphRAG~\cite{Edge2024FromLT} which uses graph indexing and community detection to generate query-focused summaries, effectively answer global questions. However, GraphRAG focuses less on the detailed steps of graph construction, such as entity resolution and relation inference. In contrast, this work places greater emphasis on the process of global KGC, including entity merging, conflict resolution, and novel triplet discovery, thereby achieving a more comprehensive and consistent knowledge representation.


\textbf{Educational Question Answering} This work also falls within the scope of applications for educational question answering. Modern NLP and Artificial Intelligence (AI) techniques have been applied to a wide range of applications, with education being a significant area. For instance, various tools have been developed focusing on writing assistance, language study, automatic grading, and quiz generation~\cite{Seyler2015GeneratingQQ, GonzlezCarrillo2021AutomaticGT, Zhang2023VISARAH,Lu2023ErrorAP,yang2023large}. Moreover, in educational scenarios, providing responses to students still requires considerable effort, as the questions often demand a high degree of relevance to the study materials and strong domain knowledge. Consequently, many studies have concentrated on developing automatic QA models~\cite{Zylich2020ExploringAQ, Hicke2023ChaTATA}, which tackle a range of queries, from logistical to knowledge-based questions. In this work, we integrate a free-text constructed KG for various QA tasks in NLP education. 


\section{Graphusion: Zero-shot Knowledge Graph Construction}

\label{sec:kgc}
We now introduce our Graphusion framework for constructing scientific KGs, shown in Fig.~\ref{fig:model}. Our approach addresses three key challenges in zero-shot KGC: 1) the input consists of free text rather than a predefined list of entities; 2) the relations encompass multiple types, and conflicts may exist among them; and 3) the output is not a single binary label but a list of triplets, making evaluation more challenging.

\textbf{Problem Definition} A $KG$ is defined as a set of triplets $KG = \{(h_i, r_i, t_i) \mid h_i, t_i \in E,\ r_i \in R,\ i = 1, 2, \ldots, n\} $,
where $E$ is the set of entities, $R$ is the set of possible relations, and $n$ is the total number of triplets in the $KG$. The task of zero-shot KGC involves taking a set of free text $T$ and generating a list of triplets $(h,r,t)$ spanning a KG. Optionally, there is an expert-annotated KG, $G_{E}$, as input, in order to provide existing knowledge. In our setting, the number of triplets of $KG$ is much larger than $G_{E}$. We select the domain to be NLP, so the entities are limited to NLP entities, with other entity types such as people, and organizations not being our focus. Referring to previous works~\cite{Luan2018MultiTaskIO}, we define 7 relations types: \texttt{Prerequisite\_of}, \texttt{Used\_for}, \texttt{Compare}, \texttt{Conjunction}, \texttt{Hyponym\_of}, \texttt{Evaluate\_for} and \texttt{Part\_of}. We will now describe the three steps of the pipeline in detail.

\subsection{Step 1: Seed Entity Generation}
Extracting domain-specific entities using LLMs under a zero-shot setting is highly challenging due to the absence of predefined entity lists. This process is not only resource-intensive but also tends to generate a large number of irrelevant entities, or entities with a bad granularity, thereby compromising the quality of extraction. To address these issues, we adopt a seed entity generation approach for efficiently extracting in-domain entities from free text~\cite{ke2024comparing}. Specifically, we utilize BERTopic~\cite{Grootendorst2022BERTopicNT} for topic modeling to identify representative entities for each topic. These representative entities serve as seed entities, denoted as $Q$. The initialized seed entities ensure high relevance in entity extraction and provide certain precision for subsequent triplet extraction.

\subsection{Step 2: Candidate Triplet Extraction}
These seed entities obtained from Step 1 would guide us to conduct entity extraction. In Step 2, we begin extracting candidate triplets from the free text. Each time, we input an entity \( q \in Q \) (\{query\}) as the query entity and retrieve documents containing this entity (\{context\}) through information retrieval. Our goal is to extract any potential triplet that includes this query entity. To achieve this, we design a Chain-of-Thought (CoT)~\cite{CoT} prompt. We first instruct the LLMs to extract in-domain entities, and then identify the possible relations between those entities and $q$. Then, we ask LLMs to discover novel triplets, even if $q$ is not initially included. This approach ensures that the seed entities play a leading role in guiding the extraction of in-domain entities. Meanwhile, the candidate triplets will encompass novel entities. We design the \textbf{Extraction Prompt} to be the following:
\vspace{4mm}
\begin{lstlisting}[language={}, captionpos=b, label={lst: zs}, basicstyle=\linespread{0.85}\ttfamily,  breaklines=true, keepspaces=true, xleftmargin=0.5pt, xrightmargin=0.5pt, numbers=none]
Given a context {context} and a query 
entity {query}, do the following: 

1. Extract the query entity and 
   in-domain entities from the context, 
   which should be fine-grained...
2. Determine the relations between 
   the query entity and the extracted 
   entities, in a triplet format: 
   (<head entity>, <relation>, <tail entity>)...
   {Relation Definition}
3. Please note some relations are 
   strictly directional...
4. You can also extract triplets from 
   the extracted entities, and the 
   query entity may not be necessary 
   in the triplets. 
\end{lstlisting}
\vspace{4mm}

After processing all the queries from the seed entity list, we save all the candidate triplets. We denote this zero-shot constructed KG by the LLM as $\mathcal{ZS-KG}$.

\subsection{Step 3: Knowledge Graph Fusion} The triplets extracted in the previous step provide a local view rather than a global perspective of each query entity. Due to the limitations of context length, achieving a global view is challenging. Additionally, the relations extracted between two entities can be conflicting, diverse, or incorrect, such as \texttt{(neural summarization methods, Used-for, abstractive summarization)} and \texttt{(neural summarization methods, Hyponym-of, abstractive summarization)}. To address the aforementioned challenge, we propose the fusion step. This approach helps reconcile conflicting relations, integrate diverse or incorrect relations effectively, and ultimately provides a global understanding of an entity pair. Specifically, for each query entity $q$, we first query from $\mathcal{ZS-KG}$, and obtain a sub-graph that contains $q$:
\begin{equation}
\text{LLM-KG} = \{ (h, r, t) \in \mathcal{ZS\text{-}KG} \mid h = q \text{ or } t = q \}.
\nonumber
\end{equation}
Optionally, if there is an expert-annotated KG $G_{E}$ available, we will also query a sub-graph, marked as $\mathcal{E-G}$. Moreover, we conduct a dynamic retrieval of $q$ again from the free text (\{background\}), to help LLMs to have a better understanding on how to resolve the conflicted triplets. This key fusion step focuses on three parts: 1)~entity merging: merge semantically similar entities, i.e., \texttt{NMT} vs \texttt{neural machine translation}; 2)~conflict resolution: for each entity pair, resolve any conflicts and choose the best one; and 3)~novel triplet inference: propose new triplet from the background text. We utilize the following \textbf{Fusion Prompt}:

\vspace{4mm}
\begin{lstlisting}[language={}, captionpos=b, label={lst: fusion}, basicstyle=\linespread{0.85}\ttfamily, breaklines=true, keepspaces=true, xleftmargin=0.5pt, xrightmargin=0.5pt, numbers=none]
Please fuse two sub-knowledge graphs 
about the entity: {entity}.

Graph 1: {LLM-KG}  
Graph 2: {E-G}

Rules for Fusing the Graphs:
1. Union the entities and edges. 
2. If two entities are similar, or 
   refer to the same entity, merge 
   them into one entity, keeping the 
   one that is meaningful or specific. 
3. Only one relation is allowed between 
   two entities. If a conflict exists, 
   read the ### Background to help 
   you keep the correct relation...
4. Once step 3 is done, consider every 
   possible entity pair not covered in
   step 2. For example, take an entity 
   from Graph 1, and match it with a 
   entity from Graph 2. Then, please
   refer to ### Background to summarize 
   new triplets. 

### Background: 
{background}

{Relation Definition}
\end{lstlisting}

\section{Experiments on Knowledge Graph Construction}

In these experiments, we investigated the general capabilities of Graphusion for KGC. 


\textbf{Dataset}
To conduct scientific KGC, we need a large-scale free-text corpus to serve as the knowledge source. We collect the proceedings papers from the ACL conference\footnote{\url{https://aclanthology.org/venues/acl/}} spanning 2017-2023, which includes a total of 4,605 valid papers. Considering that abstracts provide high-density, noise-free information and save computational resources, we perform topic modeling and KGC on the paper abstracts.

\textbf{Implementation}
We implement Graphusion on top of four settings with different LLMs: LLaMa3-70b\footnote{\url{https://llama.meta.com/llama3/}}, GPT-3.5, GPT-4 and GPT-4o. Additionally, we compare with multiple baselines, including zero-shot (GPT-4o zs) and RAG (GPT-4o RAG). We query what is the relation of a node pair as the prompt, and the zero-shot setting answers directly, while the RAG one answers with retrieved results from the same data with Graphusion. 

\textbf{Baseline}
We compare with a local graph model (GPT-4o Local), which equals to the Graphusion model without the fusion step (Step 3). Note that it is challenging to evaluate the entity quality, as a simple prompt will generate out-of-domain nodes, so we focus on comparing the relation quality. We also compare with the GraphRAG framework. Like Graphusion, we first tune the prompt with zero-shot and CoT settings for GraphRAG' entity/relation extraction. We define the relation types within the prompt. We also utilize GraphRAG's prompt auto-tuning ability to create community reports to adapt the generated knowledge graph to the NLP domain. The full manual promo tuning can be found in the Appendix.
We then feed the collected abstracts into GraphRAG to build the indexing pipelines. Specifically, we employ GPT-4o as the base LLM. In the query phase, we ask GraphRAG the relation between the given entity pairs.

\textbf{Evaluation Metrics} 
The automatic evaluation of our scientific KGC approach is challenging,  
due to the lack of ground truth graphs matching our setting. Therefore, we conduct a human evaluation of the constructed KG. For each model, we randomly sample 100 triplets and ask experts to assess both \textit{entity quality} and \textit{relation quality}, providing ratings on a scale from 1 (bad) to 3 (good). Entity quality measures the relevance and specificity of the extracted entities, while relation quality evaluates the logical accuracy of the relation between entities. We provide the annotators with the following guidelines: 

\textbf{1. Entity Quality}
\textit{Excellent (3 points)}: Both entities are highly relevant and specific to the domain. At an appropriate level of detail, neither too broad nor too specific. For example, an entity could be introduced by a lecture slide page, or a whole lecture, or possibly have a Wikipedia page. \textit{Acceptable (2 points)}: Entity is somewhat relevant, or granularity is acceptable. \textit{Poor (1 point)}: Entity is at an inappropriate level of detail, too broad or too specific.

\textbf{2. Relation Quality}
\textit{Correct (3 points)}: The relation logically and accurately describes the relation between the head and tail entities. \textit{Somewhat Correct (2 points)}: The relation is acceptable but has minor inaccuracies or there might be another better or correct answer. \textit{Incorrect (1 point)}: The relation does not logically describe the relation between the entities.









Additionally, we calculate the Inter-Annotator Agreement (IAA) between the two experts using the Kappa score.

\textbf{Results}
Tab.~\ref{tab:IAA} shows the ratings and the experts' consistency scores. Overall, the rating for entity surpasses relation, demonstrating the challenge of relation extraction. Among all the methods tested, Graphusion with GPT-4o achieves the highest performance in both entity and relation ratings. Notably, when the fusion step is omitted, performance drops significantly from 2.37 to 2.08, demonstrating the crucial role of the fusion step in enhancing relation quality within Graphusion. Notably, the performance of GraphRAG is suboptimal, partly due to modifications we made to better align it with our evaluation. Furthermore, our observations indicate that it often defaults to predicting a single relation type when uncertain (e.g., 40 \texttt{Part\_Of} relations out of 100 cases).
Additionally, the high consistency score among the experts indicates the reliability of the expert evaluation.

\begin{table}[t]
\centering
\begin{tabular}{lccccc}
\toprule
\textbf{Model} & \multicolumn{2}{c}{\textbf{Entity}} & \multicolumn{2}{c}{\textbf{Relation}} \\
\cmidrule(r){2-3} \cmidrule(r){4-5}
& \textbf{Rating} & \textbf{Kappa} & \textbf{Rating} & \textbf{Kappa} \\
\midrule
GPT-4o zs & - & - & $2.28_{\pm 0.88}$  & 0.68 \\
GPT-4o RAG & - & - & $2.28_{\pm 0.87}$  & 0.66 \\
GPT-4o Local & - & - & $2.08_{\pm 0.86}$  & 0.59 \\
GraphRAG  & - & - & $2.09_{\pm 0.70}$ & 0.56 \\
\midrule
\textit{Graphusion} & & & & \\
LLaMA  & $2.83_{\pm 0.47}$ & 0.63 & $1.82_{\pm 0.81}$ & 0.51 \\
GPT-3.5  & $2.90_{\pm 0.38}$ & 0.48 & $2.14_{\pm 0.83}$ & 0.67 \\
GPT-4  & $2.84_{\pm 0.50}$ & 0.68 & $2.36_{\pm 0.81}$ & 0.65 \\
\textbf{GPT-4o}  & $\bm{\textbf{2.92}_{\pm 0.32}}$ & 0.65 & $\bm{\textbf{2.37}_{\pm 0.82}}$ & 0.67 \\
\bottomrule
\end{tabular}
\caption{Rating for the quality of entity and relation, and IAA score for the expert evaluation.}
\label{tab:IAA}
\vspace{-5mm}
\end{table}


\begin{table*}[h]
\centering
\begin{tabular}{lcccccccc}
\toprule
\textbf{Method} & \multicolumn{2}{c}{\textbf{NLP}} & \multicolumn{2}{c}{\textbf{CV}} & \multicolumn{2}{c}{\textbf{BIO}} & \multicolumn{2}{c}{\textbf{Overall}} \\
\cmidrule(lr){2-3} \cmidrule(lr){4-5} \cmidrule(lr){6-7} \cmidrule(lr){8-9}
 & \textbf{Acc} & \textbf{F1} & \textbf{Acc} & \textbf{F1} & \textbf{Acc} & \textbf{F1} & \textbf{Acc} & \textbf{F1} \\
\midrule 
\multicolumn{9}{l}{\textit{Supervised Baselines}} \\
P2V\cite{artetxe2018emnlp} & 0.6369 & 0.5961 & 0.7642 & 0.7570 & 0.7200 & 0.7367 & 0.7070 & 0.6966 \\
BERT\cite{Devlin2019BERTPO} & 0.7088 & 0.6963 & 0.7572 & 0.7495 & 0.7067 & 0.7189 & 0.7242 & 0.7216 \\
DeepWalk\cite{Perozzi:2014:DOL:2623330.2623732} & 0.6292 & 0.5860 & 0.7988 & 0.7910 & 0.7911 & 0.8079 & 0.7397 & 0.7283 \\
Node2vec\cite{grover2016node2vec} & 0.6209 & 0.6181 & 0.8197 & 0.8172 & 0.7956 & 0.8060 & 0.7454 & 0.7471 \\
\midrule
\multicolumn{9}{l}{\textit{Zero-shot (zs)}} \\
LLaMA & 0.6058 & 0.6937 & 0.6092 & 0.6989 & 0.6261 & 0.6957 & 0.6137 & 0.6961 \\
GPT-3.5 & 0.6123 & 0.7139 & 0.6667 & 0.7271 & 0.6696 & 0.6801 & 0.6495 & 0.7070 \\
GPT-4 & 0.7639 & 0.7946 & \textbf{0.7391 }& \textbf{0.7629} & \textbf{0.7348} & \textbf{0.7737} & \textbf{0.7459} & \textbf{0.7771} \\
\midrule                            
\multicolumn{9}{l}{\textit{Zero-shot + RAG}} \\
GPT-3.5 & 0.7587 & 0.7793 & 0.6828 & 0.7123 & 0.6870 & 0.7006 & 0.7095 & 0.7307 \\
GPT-4 & \textbf{0.7755} & \textbf{0.7958} & 0.7230 & 0.7441 & 0.7174 & 0.7200 & 0.7386 & 0.7533 \\
\bottomrule
\end{tabular}
\caption{Link prediction results across all domains on the LectureBankCD test set: We present accuracy (Acc) and F1 scores. Bolded figures indicate the best performance in the zero-shot setting, while underlined scores represent the highest achievements in the supervised setting. We apply LLaMA2-70b for all for this task.}
\label{tab:main}
\vspace{-5mm}
\end{table*}

\begin{figure}[t]
    \centering
\includegraphics[width=0.47\textwidth]{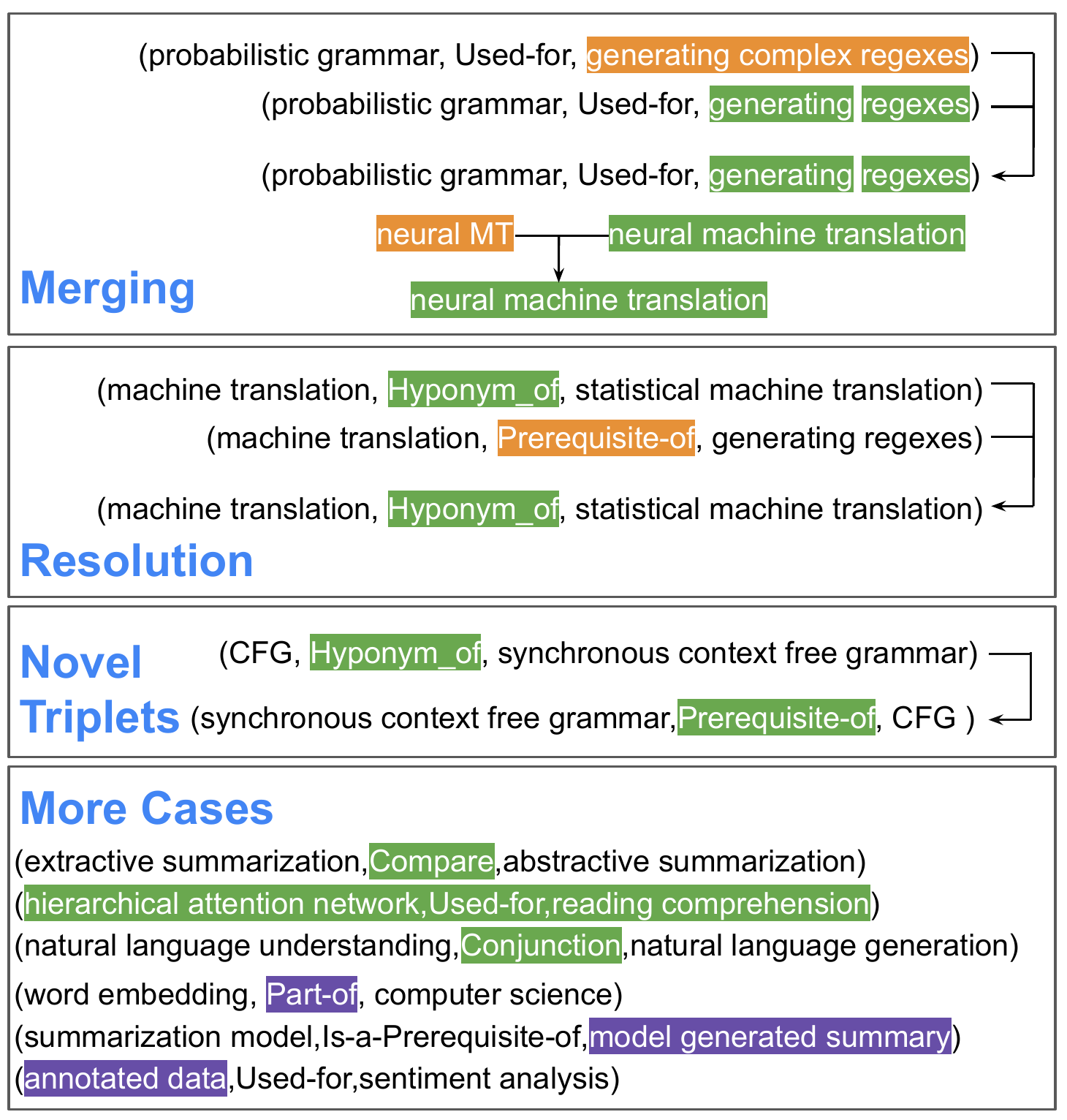}
    \caption{Case studies for Graphusion on the GPT-4o model: Correct parts are highlighted in green, resolved and merged parts in orange, and less accurate parts in purple. }
    \label{fig:graphusion_case}
    \vspace{-6mm}
\end{figure}

\textbf{Case Study: Fusion}
In Fig~\ref{fig:graphusion_case}, we present case studies from our Graphusion framework using GPT-4o. Our fusion step merges similar entities (\texttt{neural MT} and \texttt{neural machine translation}) and also resolves relational conflicts (\texttt{Prerequisite\_of} and \texttt{Hyponym\_of}). Additionally, it can infer novel triplets absent from the input. We highlight both positive and negative outputs. For instance, it correctly identifies the use of a technique for a task (\texttt{hierarchical attention network, Used\_for,\\ reading comprehension}). However, it may output less accurate triplets in entity recognition, such as entities with poor granularity (\texttt{annotated data}, \texttt{model generated summary}) and identifying very far relations (\texttt{word embedding} being categorized as part of \texttt{computer science}).

\begin{table*}[t]
\centering
\begin{tabular}{lcccc}
\toprule
\textbf{Dataset} & \textbf{Domain} & \textbf{Answer Type} & \textbf{With KG} & \textbf{Collection}\\ \midrule
CBT~\cite{Hill2015TheGP} & Open-Domain & Multiple Choice & No & Automated \\
LectureBankCD~\cite{Li2021UnsupervisedCP} &  NLP, CV, BIO & Binary & Yes & Expert-verified \\
FairytaleQA~\cite{Xu2022FantasticQA} & Open-Domain & Open-ended & No & Expert-verified \\
ChaTa~\cite{Hicke2023ChaTATA} & CS & Open-ended & No & Students \\ 
ExpertQA~\cite{Malaviya2023ExpertQAEQ} & Science & Open-ended & No & Expert-verified \\
SyllabusQA~\cite{Fernandez2024SyllabusQAAC} & Multiple & Open-ended & No &  Course syllabi\\
\midrule
TutorQA (this work) & NLP & Open-ended, Entity List, Binary & Yes & Expert-verified \\ 
\bottomrule
\end{tabular}
\caption{Comparison with other similar benchmarks: Educational or general QA benchmarks. }
\label{tab:tabqa}
\vspace{-5mm}
\end{table*}

\section{Experiments on Link Prediction}
While the task of KGC is to generate a list of triplets, including entities and their corresponding relations, we also evaluate a sub-task: focusing solely on Link Prediction (LP) for pre-defined entity pairs and a single relation type,$r=$\texttt{Prerequisite\_of}. Specifically, given an entity pair $(A, B)$, the task is to determine if a relation $r$ exists between two given entities. For instance, to learn the entity of \texttt{POS Tagging}, one must first understand the \texttt{Viterbi Algorithm}. Initially, a predefined set of entities $\calligra{C}$ is given.

We then design an \textbf{LP Prompt} to solve the task. The core part is to provide the domain name, the definition and description of the dependency relation to be predicted, and the query entities. Meanwhile, we explore whether additional information, such as entity definitions from Wikipedia and neighboring entities from training data (when available), would be beneficial. We provide detailed prompts in the Appendix.

We conduct a comprehensive evaluation on a scientific benchmark, LectureBankCD \cite{Li2021EfficientVG}, which contains entity pairs and the prerequisite labels among three domains: NLP, computer vision (CV), and bioinformatics (BIO). There are 1551, 871 and 234 entity pairs, respectively. We follow the same setting and training/testing split provided by the authors \cite{Li2021EfficientVG} and report the accuracy and F1 score, shown in Tab.~\ref{tab:main}. We compare supervised baselines, zero-shot link prediction, and zero-shot approaches using RAG models. Specifically, the RAG data predominantly consists of NLP-related content, which explains the lack of noticeable improvement in the CV and BIO domains when using RAG. Overall, the LLM method outperforms traditional supervised baselines, suggesting that LLMs have the potential to achieve higher quality in knowledge graph construction, particularly in relation prediction.

\begin{figure*}[t!]
    \centering
\includegraphics[width=1.00\textwidth]{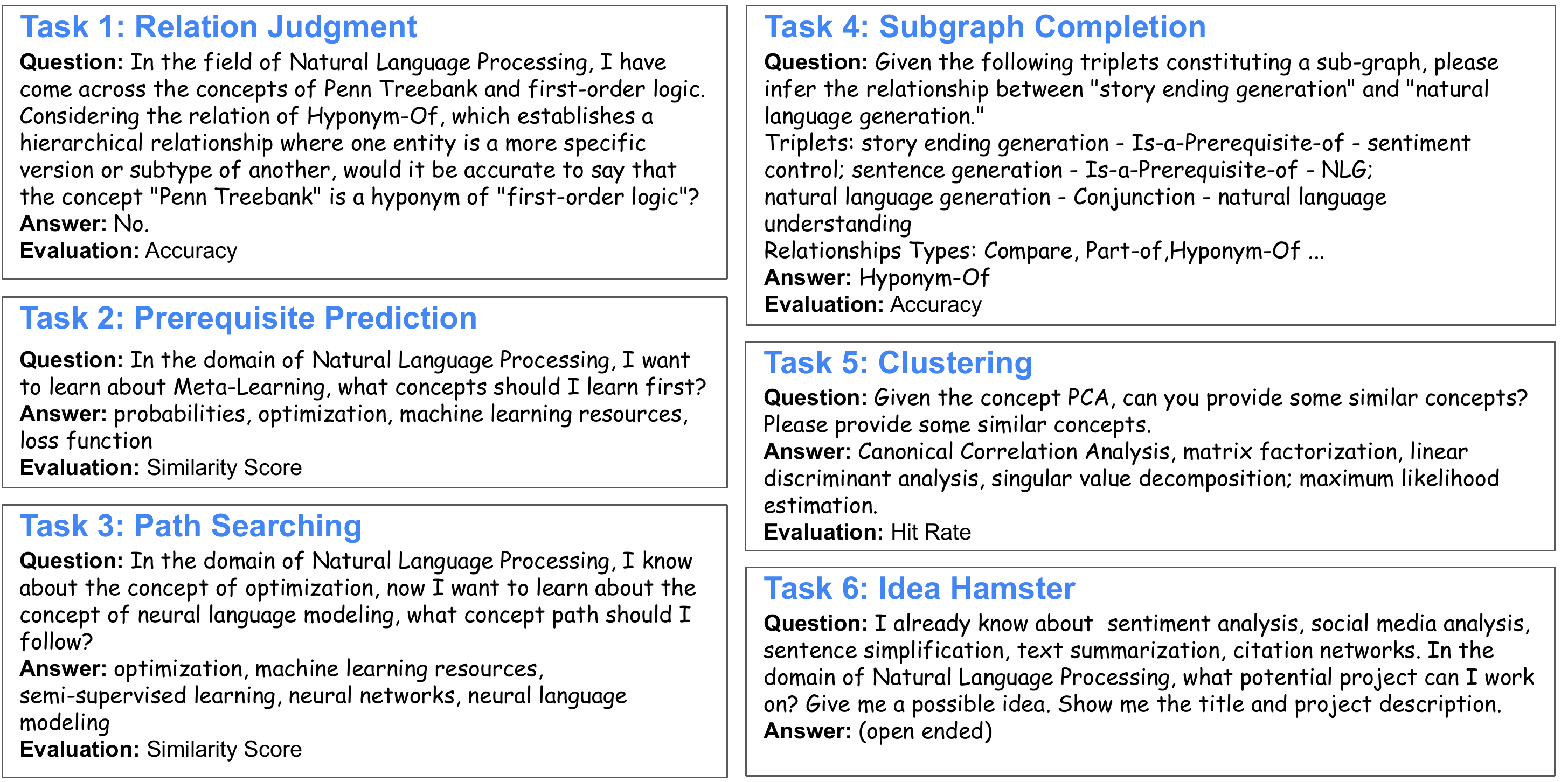}
    \caption{TutorQA tasks: We present a sample data instance and the corresponding evaluation metric for each task. Note: Task 6 involves open-ended answers, which are evaluated through human assessment. }
    \label{fig:tutorqa_sample}
    \vspace{-3mm}
\end{figure*}

\section{TutorQA: A Scientific Knowledge Graph QA Benchmark}

We aim to evaluate the practical usefulness of the Graphusion-constructed KG from an educational perspective. In NLP classes, students often have specific questions that require answers grounded in NLP domain knowledge, rather than general or logistical queries related to the course. To address this need, we introduce the TutorQA benchmark, a QA dataset designed for scientific KG QA. 

TutorQA consists of six categories, encompassing a total of 1,200 QA pairs that have been validated by human experts, simulating questions typically encountered in classes. These questions extend beyond simple syllabus inquiries, covering more complex and challenging topics that require KG reasoning, along with proficiency in text comprehension and question answering. We list some similar benchmarks in Tab~\ref{tab:tabqa}. While numerous open-domain QA benchmarks exist, our focus has been primarily on those within the scientific domain and tailored for college-level education, aligning with our objective to compare with benchmarks that can emulate a learning scenario. Among those, TutorQA is distinguished by its diversity in answer types and features expert-verified questions, ensuring a high standard of quality and relevance. 

\textbf{TutorQA Tasks}
We design different difficulty levels of the questions and divide them into 6 tasks. 
We summarize the tasks and provide example data in Fig~\ref{fig:tutorqa_sample}. More data statistics and information can be found in the supplementary materials. 

\textit{Task 1: Relation Judgment} 
The task is to assess whether a given triplet, which connects two entities with a relation, is accurate.

\textit{Task 2: Prerequisite Prediction} 
The task helps students by mapping out the key entities they need to learn first to understand a complex target topic.

\textit{Task 3: Path Searching} %
This task helps students identify a sequence of intermediary entities needed to understand a new target entity by charting a path from the graph.

\textit{Task 4: Sub-graph Completion} 
The task involves expanding the KG by identifying hidden associations between entities in a sub-graph.

\textit{Task 5: Similar Entities} 
The task requires identifying entities linked to a central idea to deepen understanding and enhance learning, aiding in the creation of interconnected curriculums.

\textit{Task 6: Idea Hamster}
The task prompts participants to develop project proposals by applying learned entities to real-world contexts, providing examples and outcomes to fuel creativity.


\textbf{Scientific Knowledge Graph Question Answering}
To address TutorQA tasks, we first utilize the Graphusion framework to construct an NLP KG. Then we design a framework for the interaction between the LLM and the graph, which includes two steps: command query and answer generation. In the command query stage, an LLM independently generates commands to query the graph upon receiving the query, thereby retrieving relevant paths. During the answer generation phase, these paths are provided to the LLM as contextual prompts, enabling it to perform QA. 

\textbf{Evaluation Metrics}
\textit{Accuracy} We report the accuracy score for Task 1 and Task 4, as they are binary classification tasks.

\textit{Similarity score} 
For Tasks 2 and 3, the references consist of a list of entities. Generally, LLMs demonstrate creativity by answering with novel entities, which are often composed of more contemporary and fresh words, even though they might not exactly match the words in the graph. Consequently, conventional evaluation metrics like keyword matching are unsuitable for these tasks. To address this, we propose the \textit{similarity score}. This metric calculates the semantic similarity between the entities in the predicted list $C_{pred}$ and the ground truth list $C_{gold}$. Specifically, as shown in Eq~\ref{eq:sim}, for an entity $m$ from the predicted list, and an entity $n$ from the ground truth list, we calculate the cosine similarity between their embeddings achieved from pre-trained BERT model~\cite{Devlin2019BERTPO}. We then average these similarity scores to obtain the similarity score.
\begin{equation}
\begin{aligned}
& Score= \frac{\sum_{m \in C_{pred}} \sum_{n \in C_{gold}} sim(m,n)}{|C_{pred}| \times |C_{gold}|}\ \ \ \ \ \ (1)
\nonumber
\label{eq:sim}
\end{aligned}
\end{equation}

By averaging the similarity scores, the final score provides a comprehensive measure of the overall semantic alignment between the predicted and ground truth entities.

\textit{Hit Rate} For Task 5, we employ the classical Hit Rate metric, expressed as a percentage. This measure exemplifies the efficiency of LLM at retrieving and presenting relevant entities in its output as compared to a provided list of target entities.

\begin{table*}[t]
\centering
\small
\begin{tabularx}{\textwidth}{lXX}
\toprule
\textbf{\textit{Question}} & \textit{I already know about \textcolor{bar_color1}{neural question answering, social media analysis, topic modeling, relation extraction, and word embedding variations}. In the domain of natural language processing, what potential project can I work on? Give me a possible idea. Show me the title and project description.} \\
\midrule
\textbf{LLaMA} & \textbf{Title: }Exploring the Impact of Emojis on Textual Emotions; \textbf{Description:} Emojis have become an integral part of online communication...The project will involve the following tasks: \textcolor{bar_color6}{ 1. Data Collection: We will collect a dataset of text messages...and their corresponding emotional labels (positive, negative, neutral). 2. Data Preprocessing: We will preprocess the data by removing stop words... }\\
\textbf{GPT-4o} & \textbf{Title:} Multi-faceted Analysis of Social Media Discussions on Climate Change; 
\textbf{Description:} This project aims to ...
Utilizing \textcolor{bar_color1}{neural question answering}, we can automatically address common inquiries people have about climate topics. \textcolor{bar_color1}{Social media analysis} will uncover trends, sentiment, and influential users shaping the conversation. \textcolor{bar_color2}{Topic modeling will identify underlying themes in these discussions} ...\\
\textbf{Ours} & \textbf{Title:} Enhanced Social Media Insights through Multilingual Aspect-Based Sentiment Analysis; \textbf{Description:}: This project aims to develop a comprehensive system for analyzing social media data to extract and understand user sentiments across multiple languages. By leveraging \textcolor{bar_color1}{neural question answering, topic modeling, and relation extraction,} the system will identify key aspects and sentiments from posts. \textcolor{bar_color3}{Dependency parsing and event extraction} will help in understanding syntactic and event-based relations... 
\textcolor{bar_color2}{Word embedding variations and multilingual word embeddings} will ensure accurate sentiment analysis across languages...\\
\bottomrule
\end{tabularx}
\caption{Case study on TutorQA Task 6: LLaMA, GPT-4o, and our pipeline with constructed KG from Graphusion (part). We highlight the \textcolor{bar_color1}{relevant entities}, \textcolor{bar_color6}{irrelevant entities},  \textcolor{bar_color2}{entity details} and \textcolor{bar_color3}{expanded relevant entities}.}
\label{tab:case_study}
\end{table*}

\textit{Expert Evaluation} In Task 6, where open-ended answers are generated without gold-standard responses, we resort to expert evaluation for comparative analysis between baseline results and our model. Despite available LLM-centric metrics like G-Eval~\cite{Liu2023GEvalNE}, the specific evaluation needs of this task warrant distinct criteria, particularly examining the persuasive and scientifically sound elements of generated project proposals. Four evaluation criteria, rated on a 1-5 scale, are employed: \textit{Entity Relevancy}: the project's alignment with the query entities. \textit{Entity Coverage}: the extent to which the project encompasses the query entities. \textit{Project Convincity}: the persuasiveness and practical feasibility of the project. \textit{Scientific Factuality}: the scientific accuracy of the information within the project. 




\begin{table}[t]
\centering
\small

\begin{subtable}[t]{\linewidth}
\centering
\begin{tabular}{lccccc}
\toprule
\textbf{Setting}     & \textbf{T1} &\textbf{ T2 }& \textbf{T3} & \textbf{T4}& \textbf{T5} \\
\midrule
GPT4o zs  & 69.20 & 64.42  & 66.61 & 44.00  & 11.45  \\   
GPT4o RAG & 64.40 & 65.06 & 69.31 & 40.80 &  10.02\\ 
Ours  & \textbf{92.00} & \textbf{80.29} & \textbf{77.85} & \textbf{50.00}  & \textbf{15.65} \\ 
\bottomrule
\end{tabular}
\caption{Evaluation on Tasks 1-5. T1, T4: accuracy; T2, T3: similarity score; T5: hit rate.}
\label{tab:tutorqa_res_a}
\end{subtable}


\begin{subtable}[t]{\linewidth}
\centering
\setlength{\tabcolsep}{2pt}
\begin{tabular}{@{}lcccc@{}}
\toprule
\textbf{ Model}     & \textbf{Relevancy} &\textbf{ Coverage }& \textbf{Convincity} & \textbf{Factuality} \\
\midrule
GPT4o zs   & 4.75 & 4.84 & 4.38 & 4.63 \\
GPT4o RAG & 4.73	& 4.71	& 4.58	& 4.71 \\
Ours & \textbf{4.85} & \textbf{4.91} & \textbf{4.72} & \textbf{4.77}  \\
\bottomrule
\end{tabular}
\caption{Expert evaluation on Task 6.}
\label{tab:tutorqa_res_b}
\end{subtable}
\caption{Results for TutorQA evaluations across various tasks. }
\label{tab:tutorqa_combined}
\vspace{-5mm}
\end{table}

\textbf{Experimental Results}
Our analysis in Tab.~\ref{tab:tutorqa_combined} compares the results based on Graphusion and two baselines, including GPT-4o zero-shot (zs) and GPT-4o with RAG (RAG). Our method with the Graphusion constructed KG shows significant improvements across Tasks~1 to~6 over the baselines. Specifically, Task~6 is evaluated by two NLP experts, with a Kappa score of 0.67, which suggests substantial agreement. The results indicate that our pipeline exhibits a marginally superior performance, particularly in the expert evaluation of \textit{Convincity} and \textit{Factuality}. This suggests that our method might be better at generating content that is not only factually accurate but also presents it in a more persuasive way to the reader. Compared to the base RAG framework, the Graphusion-generated KGs lead to better performance, particularly in Task~4 and 5, where a global understanding is essential. This improvement highlights the critical role of our core fusion step in addressing complex QA. 

\textbf{Case Study: Task 6 (Expanded relevant entities in the answer) }
To further understand how KGs could help in advanced educational scenarios, we present a case study on Task~6 in Tab.~\ref{tab:case_study}. The posed question incorporates five entities (highlighted in blue), with the task being to formulate a feasible project proposal. Although LLaMA offers a substantial project description, its content and relevance to the highlighted entities (marked in orange) are somewhat lacking. In contrast, GPT-4o not only references the queried entities but also provides detailed insights (highlighted in purple) on their potential utility within the project, such as the role of \texttt{neural question answering}. Lastly, with Graphusion constructed KG, the model provides a more comprehensive solution, elaborating on the entities and introducing additional ones (highlighted in lavender) that come from the recovered graph, like \texttt{dependency parsing} and \texttt{event extraction}, while initially addressing the queried entities.

\section{Extension on Japanese Medical Data}

In our exploration of extending Graphusion to Japanese medical data, we utilized a dataset comprising approximately 0.1 billion tokens collected from Japanese drug instructions through data crawling \cite{qiu2024towards}. Example triplets generated by Graphusion include: \begin{CJK}{UTF8}{goth}(バラシクロビル錠500mg, 抑制される, 発疹)\end{CJK} ((Valacyclovir Tablets 500mg, suppressed, rash)), \begin{CJK}{UTF8}{goth} (ゾビラックス錠400, 作用機序,ウイルスDNAの複製を阻害することによりウイルスの増殖を抑える)\end{CJK} ((Zovirax Tablets 400, mechanism of action, inhibits the replication of viral DNA to suppress the proliferation of the virus)). We randomly selected several case studies and found that the generated triplets were reasonable, demonstrating that Graphusion exhibits good generalizability. However, conducting a comprehensive evaluation is challenging due to the significant human effort required; therefore, we leave this evaluation for future work.

\section{Conclusion}


In this work, we proposed the Graphusion to construct scientific KGs from free text using LLMs. Through three key steps: seed entity generation, candidate triplet extraction, and KG Fusion, Graphusion builds KGs from a global perspective, addressing the limitations of traditional KGC methods.  Additionally, we introduced the new benchmark dataset TutorQA, which encompasses 1,200 expert-verified QA pairs across six tasks. TutorQA is specifically designed for KG-based QA in the NLP educational scenario. We developed an automated pipeline that leveraged the Graphusion-constructed KG, significantly enhancing the performance on TutorQA compared to pure LLM baselines.






\bibliographystyle{ACM-Reference-Format}
\bibliography{www25}

\appendix

\onecolumn
\section{Zero-shot Link Prediction Prompts}
\label{app:prompt}




\paragraph{LP Prompt}  
\mbox{}
\vspace{4mm}
\begin{lstlisting}[language={}, captionpos=b, label={lst: lp_1}, basicstyle=\linespread{0.85}\ttfamily,  breaklines=true, keepspaces=true, xleftmargin=0.5pt, xrightmargin=0.5pt, numbers=none]
We have two {domain} related entities: A: {entity_1} and B: {entity_2}.

Do you think learning {entity_1} will help in understanding {entity_2}?

Hints:
1. Answer YES or NO only.
2. This is a directional relation, which means if the answer is "YES", (B, A) is 
   false, but (A, B) is true.
3. Your answer will be used to create a knowledge graph.
   
{Additional Information}
\end{lstlisting}
\vspace{4mm}

\paragraph{LP Prompt With Chain-of-Thought}
\mbox{}
\vspace{4mm}
\begin{lstlisting}[language={}, captionpos=b, label={lst: lp_1}, basicstyle=\linespread{0.85}\ttfamily,  breaklines=true, keepspaces=true, xleftmargin=0.5pt, xrightmargin=0.5pt, numbers=none]
We have two {domain} related entities: A: {entity_1} and B: {entity_2}. 

Assess if learning {entity_1} is a prerequisite for understanding {entity_2}. 

Employ the Chain of Thought to detail your reasoning before giving a final answer.

# Identify the Domain and entities: Clearly define A and B within their domain. 
  Understand the specific content and scope of each entity.
  
# Analyze the Directional Relationship: Determine if knowledge of entity A is 
  essential before one can fully grasp entity B. This involves considering if A 
  provides foundational knowledge or skills required for understanding B.
  
# Evaluate Dependency: Assess whether B is dependent on A in such a way that 
  without understanding A, one cannot understand B.

# Draw a Conclusion: Based on your analysis, decide if understanding A is a 
  necessary prerequisite for understanding B.
  
# Provide a Clear Answer: After detailed reasoning, conclude with a distinct answer
  : <result>YES</result> if understanding A is a prerequisite for understanding B, 
  or <result>NO</result> if it is not.   
\end{lstlisting}
\vspace{4mm}

\clearpage
\paragraph{Extraction Prompt}
\mbox{}
\vspace{4mm}
\begin{lstlisting}[language={}, captionpos=b, label={lst: lp_1}, basicstyle=\linespread{0.85}\ttfamily,  breaklines=true, keepspaces=true, xleftmargin=0.5pt, xrightmargin=0.5pt, numbers=none]
### Instruction:
You are a domain expert in natural language processing, and now you are building a 
knowledge graph in this domain. 

Given a context (### Content), and a query entity (### entity), do the following:

1. Extract the query entity and in-domain entities from the context, which should 
   be fine-grained: could be introduced by a lecture slide page, or a whole 
   lecture, or possibly to have a Wikipedia page. 

2. Determine the relations between the query entity and the extracted entities, in 
   a triplet format: (<head entity>, <relation>, <tail entity>). The relation 
   should be functional, aiding learners in understanding the knowledge. The query 
   entity can be the head entity or tail entity. 

   We define 7 types of the relations:

   a) Compare: Represents a relation between two or more entities where a 
      comparison is being made. For example, "A is larger than B" or "X is more 
      efficient than Y."
    
   b) Part-of: Denotes a relation where one entity is a constituent or component of 
      another. For instance, "Wheel is a part of a Car."
    
   c) Conjunction: Indicates a logical or semantic relation where two or more 
      entities are connected to form a group or composite idea. For example, "Salt 
      and Pepper."
    
   d) Evaluate-for: Represents an evaluative relation where one entity is assessed 
      in the context of another. For example, "A tool is evaluated for its 
      effectiveness."
    
   e) Is-a-Prerequisite-of: This dual-purpose relation implies that one entity is 
      either a characteristic of another or a required precursor for another. For 
      instance, "The ability to code is a prerequisite of software development."
    
   f) Used-for: Denotes a functional relation where one entity is utilized in 
      accomplishing or facilitating the other. For example, "A hammer is used for 
      driving nails."
    
   g) Hyponym-Of: Establishes a hierarchical relation where one entity is a more 
      specific version or subtype of another. For instance, "A Sedan is a hyponym 
      of a Car."

3. Please note some relations are strictly directional. For example, "A tool is 
   evaluated for B" indicates (A, Evaluate-for, B), NOT (B, Evaluate-for, A). 
   Among the seven relation types, only "a) Compare" and "c) Conjunction" are not 
   direction-sensitive.

4. You can also extract triplets from the extracted entities, and the query entity 
   may not be necessary in the triplets. 
   
5. Your answer should ONLY contain a list of triplets, each triplet is in this 
   format: (entity, relation, entity). For example: "(entity, relation, entity)
   (entity, relation, entity)." No numbering and other explanations are needed. 
   
6. If ### Content is empty, output None. 
\end{lstlisting}
\vspace{4mm}

\clearpage
\paragraph{Fusion Prompt}
\mbox{}
\vspace{4mm}
\begin{lstlisting}[language={}, captionpos=b, label={lst: lp_1}, basicstyle=\linespread{0.85}\ttfamily,  breaklines=true, keepspaces=true, xleftmargin=0.5pt, xrightmargin=0.5pt, numbers=none]
### Instruction: You are a knowledge graph builder. 
    Now please fuse two sub-knowledge graphs about the entity "{entity}".

Graph 1: {LLM-KG}   Graph 2: {E-G}

Rules for Fusing the Graphs:
1. Union the entities and edges. 

2. If two entities are similar, or refer to the same entity, merge them into one 
   entity, keeping he one that is meaningful or specific. For example, "lstm" 
   versus "long short-term memory",  please keep "long short-term memory". 
   
3. Only one relation is allowed between two entities. If there is a conflict, read 
   the "### Background" to help you keep the correct relation. knowledge to keep the 
   correct one. For example, (ROUGE, Evaluate-for, question answering model) and 
   (ROUGE,Used-for , question answering model) are considered to be conflicts. 

4. Once step 3 is done, consider every possible entity pair not covered in step 2. 
   For example, take an entity from Graph 1, and match it from Graph 2. Then, 
   please refer to "### Background" to summarize new triplets.   

Hint: the relation types and their definition. You can use it to do Step 3.
We define 7 types of the relations:

   a) Compare: Represents a relation between two or more entities where a 
      comparison is being made. For example, "A is larger than B" or "X is more 
      efficient than Y."
    
   b) Part-of: Denotes a relation where one entity is a constituent or component of 
      another. For instance, "Wheel is a part of a Car."
    
   c) Conjunction: Indicates a logical or semantic relation where two or more 
      entities are connected to form a group or composite idea. For example, "Salt 
      and Pepper."
    
   d) Evaluate-for: Represents an evaluative relation where one entity is assessed 
      in the context of another. For example, "A tool is evaluated for its 
      effectiveness."
    
   e) Is-a-Prerequisite-of: This dual-purpose relation implies that one entity is 
      either a characteristic of another or a required precursor for another. For 
      instance, "The ability to code is a prerequisite of software development."
    
   f) Used-for: Denotes a functional relation where one entity is utilized in 
      accomplishing or facilitating the other. For example, "A hammer is used for 
      driving nails."
    
   g) Hyponym-Of: Establishes a hierarchical relation where one entity is a more 
      specific version or subtype of another. For instance, "A Sedan is a hyponym 
      of a Car."
   
### Background: 
{background}

### Output Instruction: 
    Output the new merged data by listing the triplets. Your answer should ONLY contain triplets in this format: (entity, relation, entity). No other explanations or numbering are needed. Only triplets, no intermediate results.
\end{lstlisting}
\vspace{4mm}



\label{app:ext_prompt}
\clearpage
\paragraph{Link Prediction with \textbf{Doc.} } 
\vspace{4mm}
\mbox{}
\begin{lstlisting}[language={}, captionpos=b, label={lst: lp_1}, basicstyle=\linespread{0.85}\ttfamily,  breaklines=true, keepspaces=true, xleftmargin=0.5pt, xrightmargin=0.5pt, numbers=none]
We have two {domain} related entities: A: {entity_1} and B: {entity_2}.

Do you think learning {entity_1} will help in understanding {entity_2}?

Hints:
1. Answer YES or NO only.
2. This is a directional relation, which means if the answer is "YES", (B, A) is 
   false, but (A, B) is true.
3. Your answer will be used to create a knowledge graph.
   
And here are related contents to help: 
{related documents concatenation}
\end{lstlisting}
\vspace{4mm}

\paragraph{Link Prediction with \textbf{Con.}}
\vspace{4mm}
\mbox{}
\begin{lstlisting}[language={}, captionpos=b, label={lst: lp_1}, basicstyle=\linespread{0.85}\ttfamily,  breaklines=true, keepspaces=true, xleftmargin=0.5pt, xrightmargin=0.5pt, numbers=none]
We have two {domain} related entities: A: {entity_1} and B: {entity_2}.

Do you think learning {entity_1} will help in understanding {entity_2}?

Hints:
1. Answer YES or NO only.
2. This is a directional relation, which means if the answer is "YES", (B, A) is 
   false, but (A, B) is true.
3. Your answer will be used to create a knowledge graph.
   
And here are related contents to help: 

We know that {entity_1} is a prerequisite of the following entities: 
{1-hop successors of entity_1 from training data};

The following entities are the prerequisites of {entity_1}:
{1-hop predecessors of entity_1 from training data}. 

We know that {entity_2} is a prerequisite of the following entities: 
{1-hop successors of entity_2 from training data};

The following entities are the prerequisites of {entity_2}:
{1-hop predecessors of entity_2 from training data}. 
\end{lstlisting}
\vspace{4mm}

\paragraph{Link Prediction with \textbf{Wiki.}} 
\vspace{4mm}
\mbox{}
\begin{lstlisting}[language={}, captionpos=b, label={lst: lp_1}, basicstyle=\linespread{0.85}\ttfamily,  breaklines=true, keepspaces=true, xleftmargin=0.5pt, xrightmargin=0.5pt, numbers=none]
We have two {domain} related entities: A: {entity_1} and B: {entity_2}.

Do you think learning {entity_1} will help in understanding {entity_2}?

Hints:
1. Answer YES or NO only.
2. This is a directional relation, which means if the answer is "YES", (B, A) is 
   false, but (A, B) is true.
3. Your answer will be used to create a knowledge graph.
   
And here are related contents to help: 
{Wikipedia introductory paragraph of {entity_1}}
{Wikipedia introductory paragraph of {entity_2}}
\end{lstlisting}
\vspace{4mm}

\clearpage
\paragraph{GraphRAG's Prompt Tuning for Entity/Relationship Extraction}
\label{app:GraphRAG's prompt}
\vspace{4mm}
\mbox{}
\begin{lstlisting}[language={}, captionpos=b, label={lst: lp_1}, 
basicstyle=\linespread{0.85}\ttfamily,  breaklines=true, keepspaces=true, xleftmargin=0.5pt, xrightmargin=0.5pt, numbers=none]

-Goal-
Given a text document that is potentially relevant to this activity, first identify all the entities needed from the text in order to capture the information and ideas in the text. Next, introduce each relation concept by defining the relation, and then report all relationships among the identified entities according to the predefined relations. These predefined relations and seed entities include:

-Relation Concepts and Definitions-:
a) Compare: Represents a relation between two or more entities where a comparison is being made. For example, "A is larger than B" or "X is more efficient than Y."
b) Part-of: Denotes a relation where one entity is a constituent or component of another. For instance, "Wheel is a part of a Car."
c) Conjunction: Indicates a logical or semantic relation where two or more entities are connected to form a group or composite idea. For example, "Salt and Pepper."
d) Evaluate-for: Represents an evaluative relation where one entity is assessed in the context of another. For example, "A tool is evaluated for its effectiveness."
e) Is-a-Prerequisite-of: This dual-purpose relation implies that one entity is either a characteristic of another or a required precursor for another. For instance, "The ability to code is a prerequisite of software development."
f) Used-for: Denotes a functional relation where one entity is utilized in accomplishing or facilitating the other. For example, "A hammer is used for driving nails."
g) Hyponym-of: Establishes a hierarchical relation where one entity is a more specific version or subtype of another. For instance, "A Sedan is a hyponym of a Car."

-Steps-
1. Identify all entities: For each identified entity, extract the following information:
- entity_name: Name of the entity,
Format each entity as ("entity"{tuple_delimiter}<entity_name>)

2. Identify all relations: From the entities identified in step 1, determine the relation between each pair of entities based on the predefined relation concepts (Compare, Part-of, Conjunction, Evaluate-for, Is-a-Prerequisite-of, Used-for, and Hyponym-of). For each pair of related entities:
- source_entity: Name of the source entity as identified in step 1
- target_entity: Name of the target entity as identified in step 1
- relationship_type: Select the appropriate relation from the predefined relations
- relationship_strength: a numeric score indicating strength of the relationship between the source entity and target entity

Format each relationship as ("relationship"{tuple_delimiter}<source_entity>{tuple_delimiter}<target_entity>{tuple_delimiter}<relationship_type>{tuple_delimiter}<relationship_strength>)
Return output: Provide the list of all entities and relationships identified in steps 1 and 2. Use {record_delimiter} as the list delimiter. When finished, output {completion_delimiter}.


######################
-Real Data-:
######################
text: {input_text}
######################
output:


\end{lstlisting}
\vspace{4mm}


\section{Graphusion Case Study: Entity Extraction}
To demonstrate the importance of the seed entity list from Step 1, we examine a selection of random entities extracted by both GraphRAG and Graphusion, as shown in Tab.~\ref{tab:entity_comparison}. All results are based on the GPT-4o backbone. GraphRAG sometimes extracts overly general terms, such as \texttt{benchmark} and \texttt{methodology}, which occur frequently in the corpus. In subsequent experiments, we will further illustrate how GraphRAG tends to extract entities with unsuitable granularity.

\begin{table}[t]
    \centering
    \begin{tabular}{cc}
        \toprule
        \textbf{GraphRAG} & \textbf{Graphusion} \\
        \midrule
        mixture-of-experts technique & code-switching tasks\\ mbart& NAS-BERT \\
        benchmark & linear indexed grammars \\
        multilingual alignment & analyzing data\\
        diffusionbert & semantic parsing\\
        proposed method & bias-variance\\
        methodology & few-shot learning\\
        \bottomrule
    \end{tabular}
    \caption{Entity comparison: Random samples from GraphRAG and Graphusion.}
    \label{tab:entity_comparison}
    \vspace{-9mm}
\end{table}

\section{Experimental Setup}
In our experimental setup, we employed Hugging Face's \texttt{LLaMA-2-70b-chat-hf}\footnote{\url{https://huggingface.co/meta-LLaMA}} and 
\texttt{LLaMA-3-70b-chat-hf}\footnote{\url{https://huggingface.co/meta-LLaMA/Meta-LLaMA-3-70B}} for LLaMA2 and LLaMA3 on a cluster equipped with 8 NVIDIA A100 GPUs. For GPT-3.5 and GPT-4, we used OpenAI's \texttt{gpt-3.5-turbo}, \texttt{gpt-4-1106-preview}, and \texttt{gpt-4o} APIs, respectively, each configured with a temperature setting of zero. The RAG models are implemented using Embedchain~\cite{embedchain}. To solve TutorQA tasks, we implemented our pipeline using LangChain\footnote{\url{https://www.langchain.com/}}. The total budget spent on this project, including the cost of the GPT API service, is approximately 500 USD.


\subsection{Additional Corpora Description}
\textbf{TutorialBank} We obtained the most recent version of TutorialBank from the authors, which consists of 15,583 manually curated resources. This collection includes papers, blog posts, textbook chapters, and other online resources. Each resource is accompanied by metadata and a publicly accessible URL. We downloaded the resources from these URLs and performed free text extraction. Given the varied data formats such as PDF, PPTX, and HTML, we encountered some challenges during text extraction. To ensure text quality, we filtered out sentences shorter than 25 words. Ultimately, this process yielded 559,217 sentences suitable for RAG and finetuning experiments.\\
\noindent\textbf{NLP-Papers} We downloaded conference papers from EMNLP, ACL, and NAACL spanning the years 2021 to 2023. Following this, we utilized Grobid (\url{https://github.com/kermitt2/grobid}) for text extraction, resulting in a collection of 4,787 documents with clean text.

\subsection{Ablation Study}

\textbf{Prompting Strategies} In Tab. ~\ref{tab: cot}, we explore the impact of different prompting strategies for entity graph recovery, comparing CoT and zero-shot prompts across both NLP and CV domains. The results indicate the introduction of CoT is not improving. We further find that CoT Prompting more frequently results in negative predictions. This finding serves as a drawback for our study, as it somewhat suppresses the performance of our system. This observation highlights the need to balance the impact of CoT on the rigor and complexity of predictions, especially in the context of graph recovery.
\begin{table}[h]
\centering
\begin{tabular}{lcccc}
\toprule
  Model          & \multicolumn{2}{c}{NLP}  &  \multicolumn{2}{c}{CV} \\ 
\cmidrule(r){2-3} \cmidrule(lr){4-5} 
     & Acc & F1  & Acc & F1  \\
\midrule
GPT-4 zs     & 0.7639 & 0.7946 & 0.7391&	0.7629 \\
GPT-4 CoT  & 0.7342 & 0.6537 &  0.6122 &0.4159 \\
\bottomrule
\end{tabular}
\caption{Comparison of zero-shot and CoT prompts with GPT-4: Results on NLP and CV. }
\label{tab: cot}
\end{table}

\noindent\textbf{Finetuning} We further explore the impact of finetuning on additional datasets, with results detailed in Table~\ref{tab:finetuning}. Specifically, we utilize LLaMA2-70b~\cite{touvron2023llama}, finetuning it on two previously mentioned datasets: TutorialBank and NLP-Papers. Both the zero-shot LLaMA and the finetuned models are employed to generate answers. As these answers are binary (\texttt{YES} or \texttt{NO}), we can calculate both the accuracy and F1 score for evaluation. However, the results indicate that finetuning does not yield positive outcomes. This can be attributed to two potential factors: 1) the poor quality of data, and 2) limited effectiveness in aiding the graph recovery task. We leave this part as the future work. 

\begin{table*}[h]
\centering
\begin{tabular}{lcc}
\toprule
\textbf{Dataset} & \textbf{Acc} & \textbf{F1} \\
\midrule
LLaMA2-70b & \textbf{0.6058} & \textbf{0.6937} \\	
TutorialBank & 0.4739  & 0.0764 \\
NLP Papers & 0.5435 & 0.6363 \\
\bottomrule
\end{tabular}
\caption{Comparison of the effect of finetuning: Results on NLP domain.}
\label{tab:finetuning}
\end{table*}

\subsection{Ablation Study: RAG Data for Link Prediction} We explore the potential of external data in enhancing entity graph recovery. This is achieved by expanding the \texttt{\{Additional Information\}} part in the \textbf{LP Prompt}. We utilize LLaMa as the \textbf{Base} model, focusing on the NLP domain. We introduce three distinct settings: \textbf{Doc.}: In-domain lecture slides data as free-text; \textbf{Con.}: Adding one-hop neighboring entities from the training set as additional information related to the query entities. \textbf{Wiki.}: Incorporating the introductory paragraph of the Wikipedia page of each query entity. As illustrated in Fig \ref{fig:info_comp}, our findings indicate that incorporating LectureBankCD documents (Doc.) significantly diminishes performance. This decline can be attributed to the introduction of noise and excessively lengthy content, which proves challenging for the LLM to process effectively. Conversely, the inclusion of neighboring entities (Con.) markedly enhances the base model's performance. However, it relies on training data, rendering it incompatible with our primary focus on the zero-shot setting. Incorporating Wikipedia content also yields improvements and outperforms the use of LectureBankCD, likely due to higher text quality. 

\begin{figure}[ht]
    \centering
    
\begin{tikzpicture}
\centering
\begin{axis}[
    ybar,
    bar width=6pt,
    enlarge x limits=0.45,
    legend style={
        at={(-0.2,0.5)}, 
        anchor=east,     
        legend columns=1 
    },
    symbolic x coords={Acc, F1},
    xtick=data,
    height=3cm,
    width=6cm,
]

\addplot[fill=color1, draw opacity=0, draw=none] coordinates {(Acc,0.6058) (F1,0.6937)};
\addlegendentry{Base}

\addplot[fill=color2, draw=none, draw opacity=0] coordinates {(Acc,0.5742) (F1,0.6783) };
\addlegendentry{Doc.}

\addplot[fill=color3, draw=none, draw opacity=0] coordinates {(Acc,0.6548) (F1,0.7187) };
\addlegendentry{Con.}

\addplot[fill=color4, draw=none, draw opacity=0] coordinates {(Acc,0.6503) (F1,0.7133) };
\addlegendentry{Wiki.}
\end{axis}
\end{tikzpicture}

\caption{Link Prediction Ablation Study: Comparison of models with external data.}
\label{fig:info_comp}

\end{figure}

\section{TutorQA}
\label{app:tutorqa_info}

\subsection{Benchmark Details}
We show the data analysis in Tab.~\ref{tab:data_stats}. 

\begin{table}[h]
\centering
\begin{tabular}{cccccccc}
\toprule
\textbf{Task} & \multicolumn{3}{c}{\textbf{Question Token}} & \multicolumn{3}{c}{\textbf{entity Count}}  & \textbf{Number}\\ 
\cmidrule(lr){2-4} \cmidrule(lr){5-7} 
 & \textbf{Max} & \textbf{Min} & \textbf{Mean} & \textbf{Max} & \textbf{Min} & \textbf{Mean} &  \\ \midrule
T1 & 77 & 61 & 68.00 & - & - & - & 250\\ 
T2 & 27 & 22 & 23.48 & 7 & 1 & 1.79  & 250\\ 
T3 & 40 & 34 & 36.66 & 8 & 2 & 3.36 & 250\\ 
T4 & 88 & 76 & 83.00 & - & - & - & 250\\ 
T5 & 21 & 18 & 19.26 & 8 & 1 & 4.76 & 100\\ 
T6 & 54 & 42 & 48.62 & - & - & - & 100\\ \bottomrule
\end{tabular}
\caption{TutorQA data statistics comparison: The answers in T1 are only "True" or "False", and the answers in T4 are relations, while the answers in T6 are free text with open-ended answers.}
\label{tab:data_stats}
\end{table}

\subsection{GraphRAG Results}
\label{app:graphrag_tutorqa}

We extend the results in Tab.~\ref{tab:tutorqa_combined} by adding GraphRAG as a baseline, the full version of the evaluation is shown in Tab.~\ref{tab:tutorqa_combined2}. 
Based on the established indexing pipelines in knowledge graph construction, we utilize GraphRAG's query engine with the local search method to directly ask the questions in TutorQA. Notably, the performance of GraphRAG appears less satisfactory, which may be due to an evaluation approach that is not well-suited for GraphRAG's results. For example, in Task 5, GraphRAG produces concepts with very broad or specific terms with a bad granularity, such as \textit{predict sentiment, emotion cause pair extraction, emotional support conversation}. This observation holds across other tasks, where achieving higher scores requires a more granular concept list. This indicates the critical importance of Step 1, which involves generating a well-defined seed concept, in the Graphusion pipeline. 

\begin{table}[t]
\centering

\begin{subtable}[t]{\linewidth}
\centering
\begin{tabular}{lccccc}
\toprule
\textbf{Setting}     & \textbf{T1} &\textbf{ T2 }& \textbf{T3} & \textbf{T4}& \textbf{T5} \\
\midrule
GPT4o zs  & 69.20 & 64.42  & 66.61 & 44.00  & 11.45  \\   
GPT4o RAG & 64.40 & 65.06 & 69.31 & 40.80 &  10.02\\ 
GraphRAG & 60.40 & 64.19 & 67.45 & 42.00 & 8.96\\
Ours  & \textbf{92.00} & \textbf{80.29} & \textbf{77.85} & \textbf{50.00}  & \textbf{15.65} \\ 
\bottomrule
\end{tabular}
\caption{Evaluation on Tasks 1-5. T1, T4: accuracy; T2, T3: similarity score; T5: hit rate.}
\label{tab:tutorqa_res_a}
\end{subtable}

\vspace{4mm} 

\begin{subtable}[t]{\linewidth}
\centering
\setlength{\tabcolsep}{2pt}
\begin{tabular}{@{}lcccc@{}}
\toprule
\textbf{ Model}     & \textbf{Relevancy} &\textbf{ Coverage }& \textbf{Convincity} & \textbf{Factuality} \\
\midrule
GPT4o zs   & 4.75 & 4.84 & 4.38 & 4.63 \\
GPT4o RAG & 4.73	& 4.71	& 4.58	& 4.71 \\
GraphRAG & 3.94 & 4.08 & 4.13 & 4.45\\
Ours & \textbf{4.85} & \textbf{4.91} & \textbf{4.72} & \textbf{4.77}  \\
\bottomrule
\end{tabular}
\caption{Expert evaluation on Task 6.}
\label{tab:tutorqa_res_b}
\end{subtable}
			
\caption{Results for TutorQA evaluations across various tasks. }
\label{tab:tutorqa_combined2}
\end{table}

\subsection{Task 2 and Task 3: case study}

\textbf{Entity counts} As depicted in Fig ~\ref{fig:task2,3}, we evaluate the average number of entities created by GPT-4o zs and our Graphusion framework in the responses for Task~2 and Task~3, in which both tasks require the model to give a list of reasonable entities. The results show that without the enhancement of KG retrieved information, GPT-4o tends to mention more entities in the generated response (Task 2: 11.04, Task 3: 11.54), which might be irrelevant or broad. 

\begin{figure}[t]
    \centering
        \begin{tikzpicture}
        \small
        \begin{axis}[
            ybar,
            bar width=15pt, 
            enlarge x limits=0.5, 
            legend style={
                at={(0.05,0.95)}, 
                anchor=north west,
                legend columns=2,
            },
            ymax = 35,
            symbolic x coords={Task 2, Task 3},
            xtick=data,
            height=3cm, 
            width = 7cm,
            nodes near coords, 
            nodes near coords align={vertical},
        ]
        \addplot[fill=color1, draw=none, draw opacity=0] coordinates {(Task 2, 2.84) (Task 3,2.87)};
        \addlegendentry{Graphusion}
        \addplot[fill=color2, draw=none, draw opacity=0] coordinates {(Task 2, 11.04) (Task 3, 11.54)};
        \addlegendentry{Zero-shot}
        \end{axis}
        \end{tikzpicture}
    \caption{Entity counts in Task 2 and Task 3.}
    \label{fig:task2,3}
\end{figure}

\begin{table*}[h]
\centering
\begin{tabularx}{\textwidth}{lX}
\toprule
\textbf{\textit{Question} [Task2]} & \textit{In the field of Natural Language Processing, I want to learn about \textbf{multilingual model}. What entities should I learn first?} \\
\midrule
\textbf{GPT-4o} & \textbf{Tokenization, Embeddings, Transfer Learning, Cross-lingual Transfer, Zero-shot Learning, Multilingual Corpora, Language Modeling, Fine-tuning, Evaluation Metrics, Pretrained Models} \\
\textbf{Ours} & \textbf{language models, machine translation, cross-lingual embeddings, transfer learning, tokenization, fine-tuning} \\
\midrule
\textbf{\textit{Question} [Task3]} & \textit{In the field of Natural Language Processing, I know about \textbf{natural language processing intro}, now I want to learn about \textbf{t-sne}. What entity path should I follow?}\\
\midrule
\textbf{GPT-4o} & \textbf{natural language processing, dimensionality reduction, t-SNE, perplexity, high-dimensional data, data visualization, machine learning} \\
\textbf{Ours} & \textbf{natural language processing intro, vector representations, t-sne} \\
\bottomrule
\end{tabularx}
\caption{Case study on TutorQA Task 2 and Task 3: GPT-4o, and GPT-4o-Graphusion.}
\label{tab:case_study_task2,3}

\end{table*}

\begin{table*}[h]
\centering
\begin{tabularx}{\textwidth}{lX}
\toprule
\textbf{\textit{Question}} & \textit{Given the following edges constituting an entity subgraph, please identify and select the possible type of relationship between \textbf{natural language generation} and \textbf{natural language understanding}.} \\

\midrule
\textbf{GPT-4o} & \textbf{Is-a-Prerequisite-of} \\
\textbf{Ours} & \textbf{Conjunction} \\

\bottomrule
\end{tabularx}
\caption{Case study on TutorQA Task 4: GPT-4o, and GPT-4o-Graphusion.}
\label{tab:case_study_task4}
\end{table*}

\section{Knowledge Graph Construction Analysis}

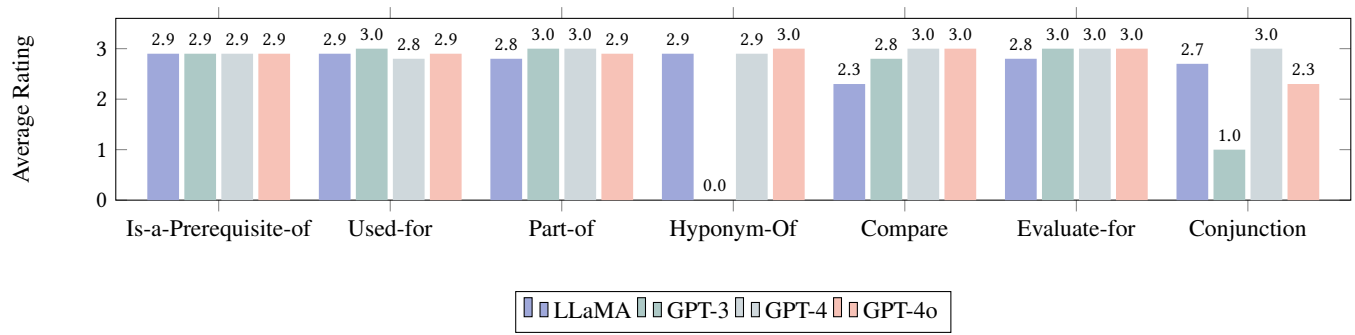
\begin{figure*}[h]
    \centering
    \begin{adjustbox}{center}
    \begin{tikzpicture}
    \centering
    \begin{axis}[
        ybar,
        bar width=12pt,
        enlarge x limits=0.1,
        enlarge y limits={value=0.2,upper},
        legend style={
            at={(0.5,-0.5)}, 
            anchor=north,    
            legend columns=4 
        },
        ylabel={Average Rating},
        symbolic x coords={Is-a-Prerequisite-of, Used-for, Part-of, Hyponym-Of, Compare, Evaluate-for, Conjunction},
        xtick=data,
        height=4cm,
        width=18cm,
        nodes near coords,
        nodes near coords align={vertical},
        every node near coord/.append style={
            font=\footnotesize,
            /pgf/number format/.cd,
            fixed,
            precision=1,
            fixed zerofill
        }
    ]

    \addplot[fill=color1, draw=none] coordinates {(Is-a-Prerequisite-of, 2.9) (Used-for, 2.9) (Part-of, 2.8) (Hyponym-Of, 2.9) (Compare, 2.3) (Evaluate-for, 2.8) (Conjunction, 2.7)};
    \addlegendentry{LLaMA}

    \addplot[fill=color2, draw=none] coordinates {(Is-a-Prerequisite-of, 2.9) (Used-for, 3.0) (Part-of, 3.0) (Hyponym-Of, 0.0) (Compare, 2.8) (Evaluate-for, 3.0) (Conjunction, 1.0)};
    \addlegendentry{GPT-3}

    \addplot[fill=color3, draw=none] coordinates {(Is-a-Prerequisite-of, 2.9) (Used-for, 2.8) (Part-of, 3.0) (Hyponym-Of, 2.9) (Compare, 3.0) (Evaluate-for, 3.0) (Conjunction, 3.0)};
    \addlegendentry{GPT-4}

    \addplot[fill=color4, draw=none] coordinates {(Is-a-Prerequisite-of, 2.9) (Used-for, 2.9) (Part-of, 2.9) (Hyponym-Of, 3.0) (Compare, 3.0) (Evaluate-for, 3.0) (Conjunction, 2.3)};
    \addlegendentry{GPT-4o}

    \end{axis}
    \end{tikzpicture}
    \end{adjustbox}
    \caption{Entity quality rating by human evaluation, grouped by relation type.}
    \label{fig:entity_quality}
\end{figure*}

\textbf{Average Rating} We compare expert ratings on the Graphusion KGC results produced by four models: LLaMA, GPT-3.5, GPT-4, and GPT-4o. Fig.~\ref{fig:entity_quality} and \ref{fig:relation_quality} display the average ratings for entity quality and relation quality, respectively, grouped by relation type. Most types achieve an average rating of around 3 (full score) in entity quality, indicating that the extracted triplets contain good in-domain entities. In contrast, the ratings for relation quality are slightly lower. GPT-4 and GPT-4o perform better in relation prediction.

\begin{figure*}[h]
    \centering
    \begin{adjustbox}{center}
    \begin{tikzpicture}
    \centering
    \begin{axis}[
        ybar,
        bar width=12pt,
        enlarge x limits=0.1,
        enlarge y limits={value=0.2,upper},
        legend style={
            at={(0.5,-0.5)}, 
            anchor=north,    
            legend columns=4 
        },
        ylabel={Average Rating},
        symbolic x coords={Is-a-Prerequisite-of, Used-for, Part-of, Hyponym-Of, Compare, Evaluate-for, Conjunction},
        xtick=data,
        height=4cm,
        width=18cm,
        nodes near coords,
        nodes near coords align={vertical},
        every node near coord/.append style={
            font=\footnotesize,
            /pgf/number format/.cd,
            fixed,
            precision=1,
            fixed zerofill
        }
    ]

    \addplot[fill=color1, draw=none] coordinates {(Is-a-Prerequisite-of, 2.9) (Used-for, 2.9) (Part-of, 2.8) (Hyponym-Of, 2.9) (Compare, 2.3) (Evaluate-for, 2.8) (Conjunction, 2.7)};
    \addlegendentry{LLaMA}

    \addplot[fill=color2, draw=none] coordinates {(Is-a-Prerequisite-of, 2.9) (Used-for, 3.0) (Part-of, 3.0) (Hyponym-Of, 0.0) (Compare, 2.8) (Evaluate-for, 3.0) (Conjunction, 1.0)};
    \addlegendentry{GPT-3}

    \addplot[fill=color3, draw=none] coordinates {(Is-a-Prerequisite-of, 2.9) (Used-for, 2.8) (Part-of, 3.0) (Hyponym-Of, 2.9) (Compare, 3.0) (Evaluate-for, 3.0) (Conjunction, 3.0)};
    \addlegendentry{GPT-4}

    \addplot[fill=color4, draw=none] coordinates {(Is-a-Prerequisite-of, 2.9) (Used-for, 2.9) (Part-of, 2.9) (Hyponym-Of, 3.0) (Compare, 3.0) (Evaluate-for, 3.0) (Conjunction, 2.3)};
    \addlegendentry{GPT-4o}

    \end{axis}
    \end{tikzpicture}
    \end{adjustbox}
    \caption{Relation quality rating by human evaluation, grouped by relation type.}
    \label{fig:relation_quality}
\end{figure*}



\begin{figure*}[h]
    \centering
    \begin{tikzpicture}
        \pie[color={color1,color2,color3,color4,color5,color6,pie_rose},
             text=,
             radius=3.3,
             before number=\phantom{0},
             after number={\%\scriptsize},
             explode=0.1]{
            33/Is-a-Prerequisite-of,
            13/Used-for,
            19/Part-of,
            10/Compare,
            9/Evaluate-for,
            7/Conjunction,
            9/Hyponym-Of
        }
        \node at (0, -4) {LLaMA};
    \end{tikzpicture}
    \hspace{0.7cm}
    \begin{tikzpicture}
        \pie[color={color1,color2,color3,color4,color5,color6,pie_rose},
             text=,
             radius=3.3,
             before number=\phantom{0},
             after number={\%\scriptsize},
             explode=0.1]{
            45/Is-a-Prerequisite-of,
            18/Used-for,
            4/Part-of,
            18/Compare,
            14/Evaluate-for,
            1/Conjunction
        }
        \node at (0, -4) {GPT-3};
    \end{tikzpicture}
    \hspace{0.7cm}
    \begin{tikzpicture}
        \pie[color={color1,color2,color3,color4,color5,color6,pie_rose},
             text=,
             radius=3.3,
             before number=\phantom{0},
             after number={\%\scriptsize},
             explode=0.1]{
            37/Is-a-Prerequisite-of,
            38/Used-for,
            8/Part-of,
            5/Compare,
            3/Evaluate-for,
            1/Conjunction,
            8/Hyponym-Of
        }
        \node at (0, -4) {GPT-4};
    \end{tikzpicture}
    \hspace{0.5cm}
    \begin{tikzpicture}
        \pie[color={color1,color2,color3,color4,color5,color6,pie_rose},
             text=,
             radius=3.3,
             before number=\phantom{0},
             after number={\%\scriptsize},
             explode=0.1]{
            25/Is-a-Prerequisite-of,
            33/Used-for,
            13/Part-of,
            4/Compare,
            9/Evaluate-for,
            3/Conjunction,
            13/Hyponym-Of
        }
        \node at (0, -4) {GPT-4o};
    \end{tikzpicture}
    \begin{tikzpicture}
        \node at (0, 0) {\begin{tabular}{cccc}
            \textcolor{color1}{\rule{1cm}{0.4cm}} & Is-a-Prerequisite-of &
            \textcolor{color2}{\rule{1cm}{0.4cm}} & Used-for \\
            \textcolor{color3}{\rule{1cm}{0.4cm}} & Part-of &
            \textcolor{color4}{\rule{1cm}{0.4cm}} & Compare \\
            \textcolor{color5}{\rule{1cm}{0.4cm}} & Evaluate-for &
            \textcolor{color6}{\rule{1cm}{0.4cm}} & Conjunction \\
            \textcolor{pie_rose}{\rule{1cm}{0.4cm}} & Hyponym-Of & & \\
        \end{tabular}};
    \end{tikzpicture}
    \caption{Relation type distribution.}
    \label{fig:relation type distribution}
\end{figure*}
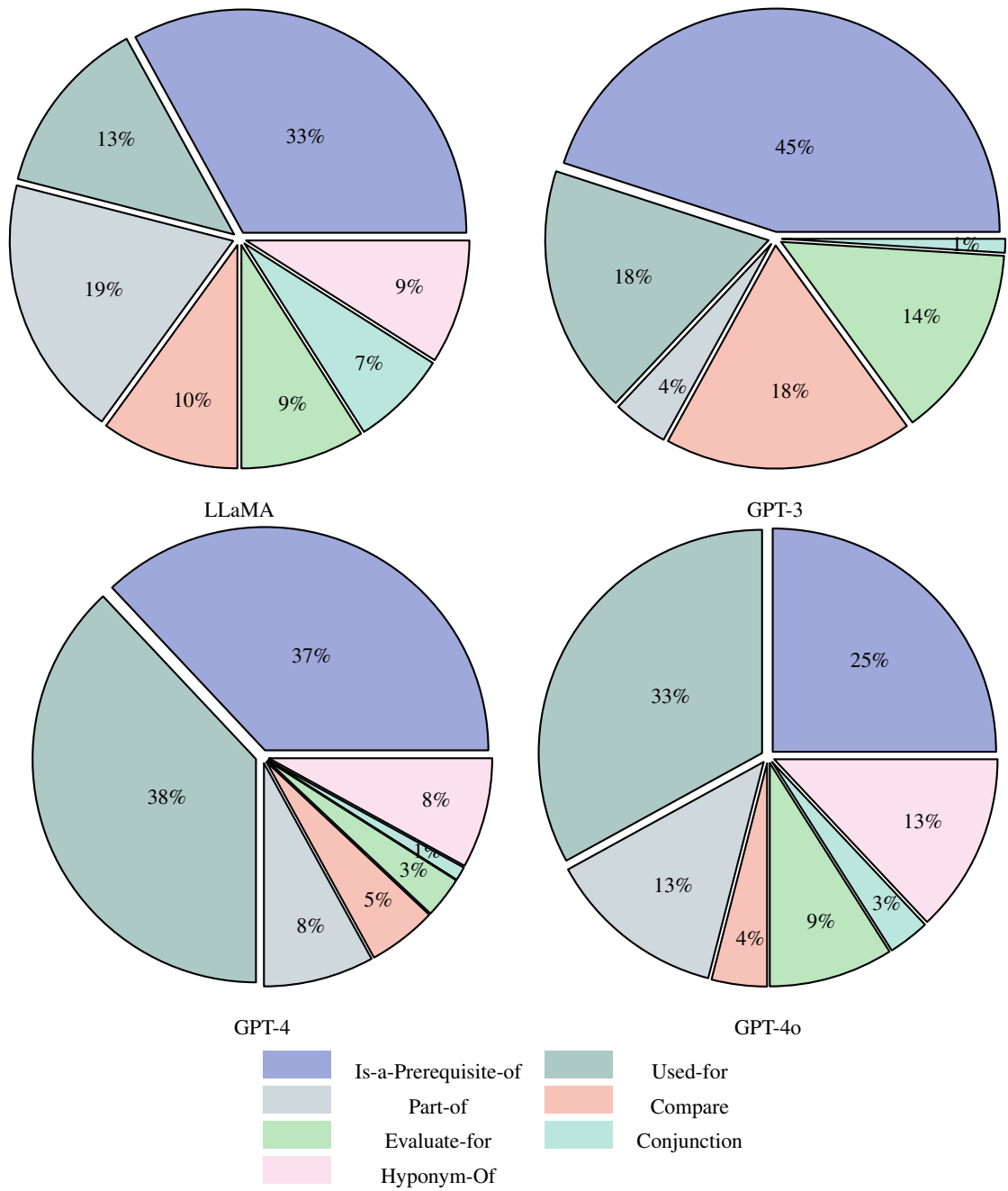

\noindent\textbf{Relation Type Distribution}  
We then compare the Graphusion results for each relation type across the four selected base LLMs, as shown in Fig.~\ref{fig:relation type distribution}. All models tend to predict \texttt{Prerequisite\_of} and \texttt{Used\_For} relations. The results from LLaMA show relatively even distributions across relation types, whereas the results from the GPT family do not.

\begin{figure}[h]
    \centering
\includegraphics[width=1\textwidth]{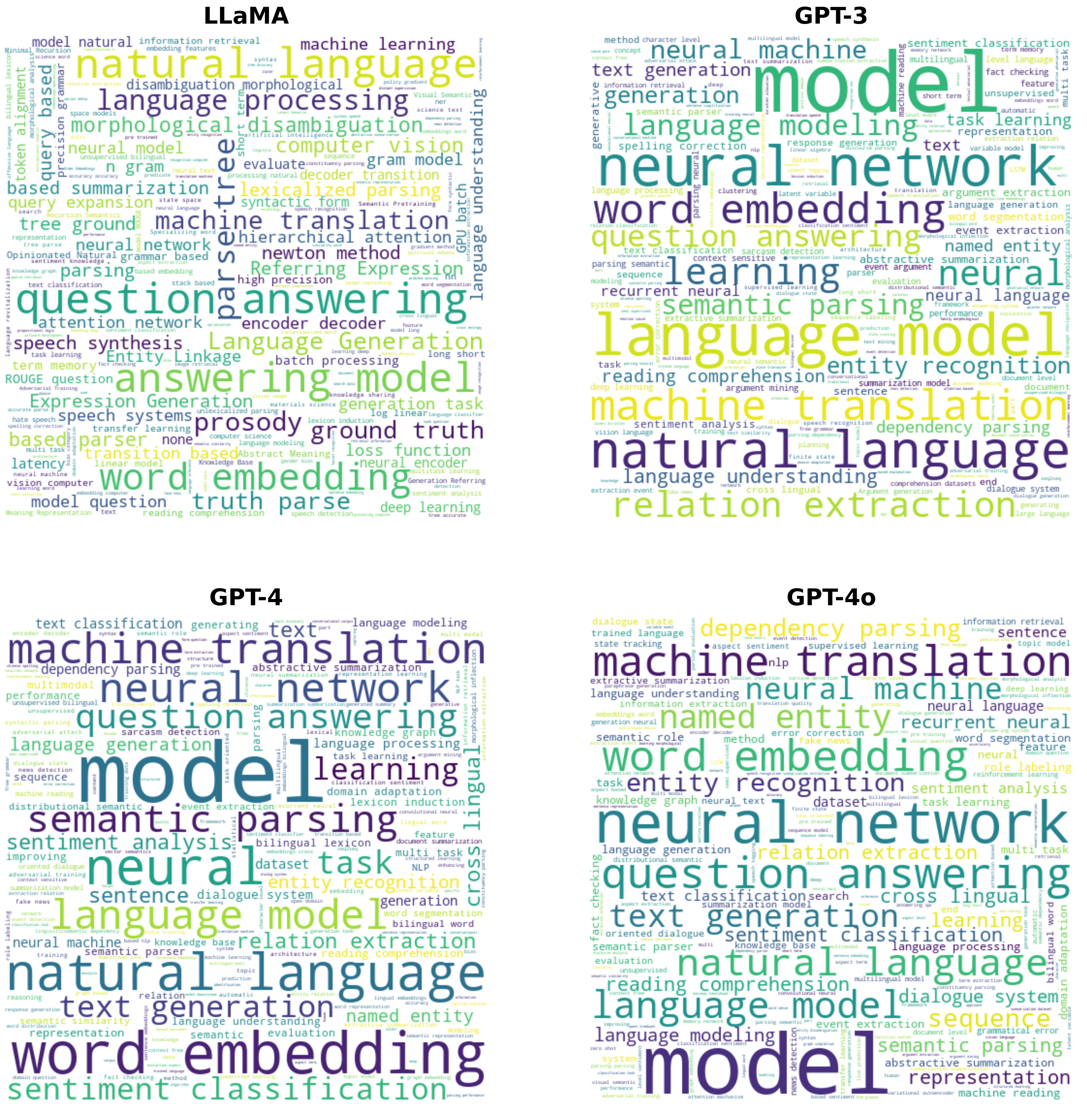}

    \caption{Word cloud visualization for extracted entities.}
    \label{fig:word_cloud}

\end{figure}

\noindent\textbf{Word cloud Visualization} Finally, in Fig.~\ref{fig:word_cloud}, we present a word cloud visualization of the entities extracted by Graphusion, comparing the four base LLMs. High-frequency entities include \texttt{word embedding}, \texttt{model}, \texttt{neural network}, \texttt{language model}, and others.

\end{document}